\documentclass[10pt,twocolumn,letterpaper]{article}

\usepackage[pagenumbers]{cvpr} 

\usepackage{algorithm}
\usepackage{algpseudocode}

\usepackage{multirow}

\usepackage{subcaption}

\def\makinhome{/home/makin}


\ifx\homepath\makinhome
    \newcommand{\stydir}{../../stys}
    
    \newcommand{\tikzdir}{../../tikzpics/ACtx}
    \newcommand{\pngdir}{../../../jpgs/ACtx}
\else
    \newcommand{\stydir}{stys}
    
    \newcommand{\tikzdir}{tikzpics}
    
    \newcommand{\pngdir}{pngs}
\fi

\providecommand{\stydir}{../stys}

\providecommand{\tikzdir}{../tikzpics}

\usepackage{%
	amsmath,amsfonts,amssymb,
	graphicx,rotating,booktabs,
	versions,ifthen,siunitx,bbm,
}
\usepackage[dvipsnames]{xcolor}
\usepackage{pgf,tikz,pgfplots} 
\usetikzlibrary{%
 	positioning,calc,patterns,shapes,arrows,intersections,backgrounds,fit,shadings,decorations.text,shadows,fadings,bayesnet,arrows.meta,spy
}
\usepgfplotslibrary{colorbrewer,colormaps,fillbetween,groupplots}

\ifdefined\matttlderiv
\else
	\makeatletter
	\input{\stydir/jgmstd.sty}
	\makeatother
\fi

\newboolean{INCLFIGS} 
\setboolean{INCLFIGS}{true}

\usepackage{subcaption}
\newcommand{\captioning}[2]{\caption{{\bf #1} {#2}}}
\ifthenelse{\boolean{INCLFIGS}}{
	\captionsetup[subfloat]{
		font={Large,sf},
		labelformat=simple,
		labelsep=none}
}{
	\captionsetup[subfloat]{labelformat=empty,labelsep=none}
}

\newcommand{\colorprovide}[2]{\@ifundefinedcolor{#1}{\colorlet{#1}{#2}}{}}

\usepackage{\stydir/distributions}

\renewcommand{\captioning}[2]{\caption{#1 {#2}}}
\renewcommand{\eqn}[1]{Eq.\ (\ref{eqn:#1})}

\newcommand{\rv}[1]{\mathbf{#1}}

\newcommand{\schematic}{
\begin{figure}[t]
  \centering
    \includegraphics[width=1.0\linewidth]{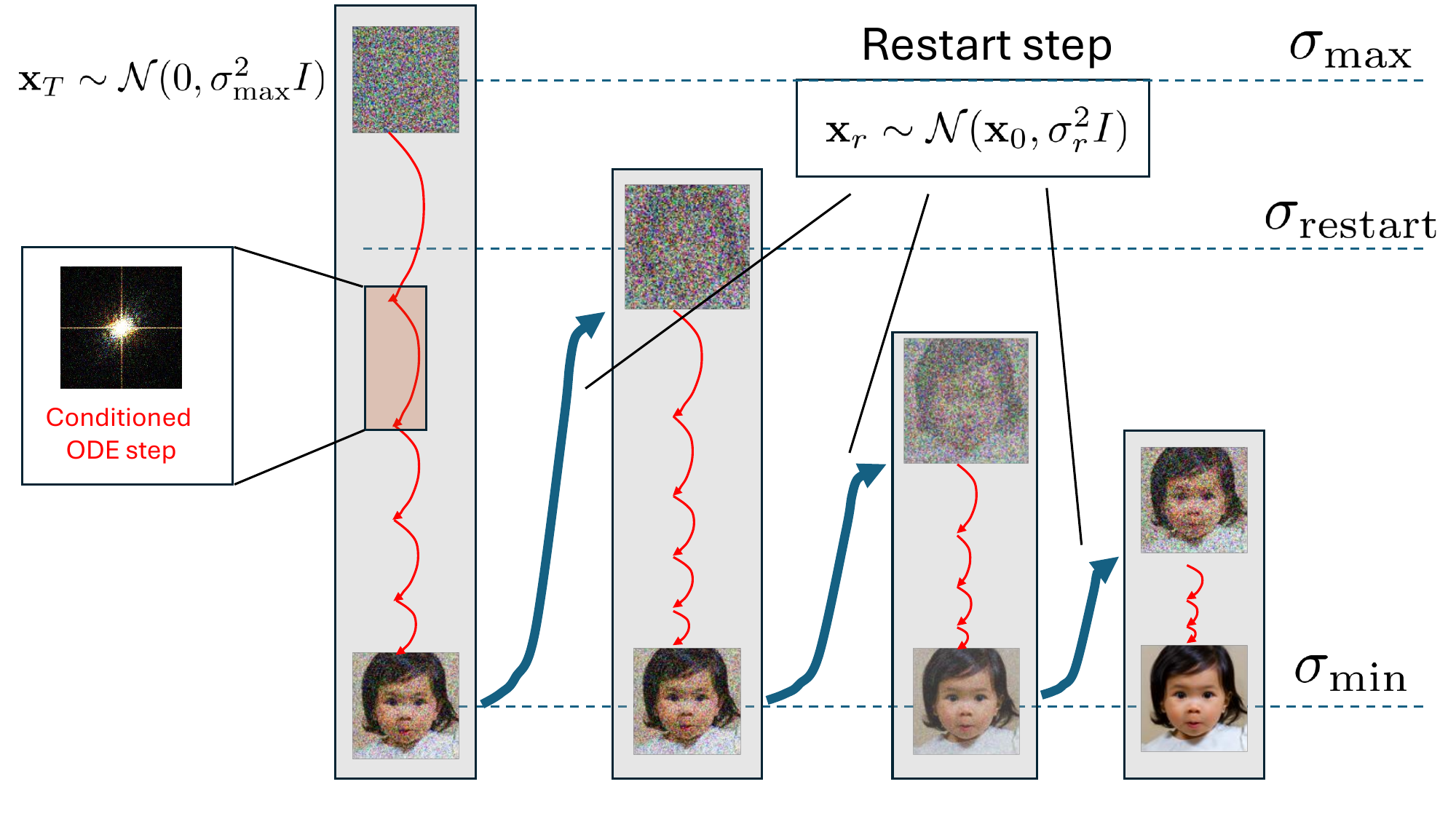}
    \captioning{Schematic conceptually demonstrating RePS.}{The process begins by sampling from $\nrml{\vect{0}}{\sigma(T)^2\mat{I}}$ and solving the measurement-conditioned ODE (steps indicated by red arcs). Once the sample reaches the noise level corresponding to $\sigma(0)$, the sampler is ``restarted'' by adding noise that returns it to $\sigma(r)$. This procedure is repeated while annealing down the restart noise level.}

    \label{fig:schematic}
\end{figure}
}

\newcommand{\qualitative}{
\begin{figure*}[t]
  \centering
    \newcommand{\figwidth}{0.16\linewidth}%
    \newcommand{\marksize}{2.5}%
    \setlength{\tabcolsep}{0pt}%
    \newcommand{\DATASET}{ffhq}
    \newcommand{\subdir}{\pngdir/\DATASET}%
    \begin{tabular}{@{}ccccccccccc@{}}
         \multicolumn{3}{c}{Nonlinear} & \multicolumn{3}{c}{Linear} \\    

        \includegraphics[width=\figwidth]{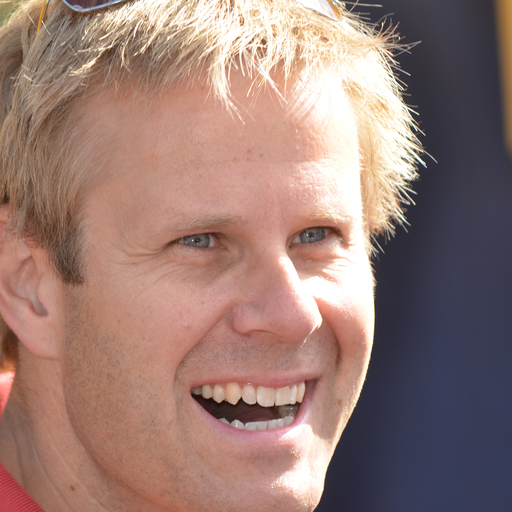} &
        \includegraphics[width=\figwidth]{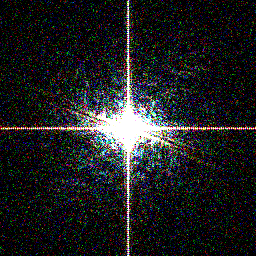} &
        \includegraphics[width=\figwidth]{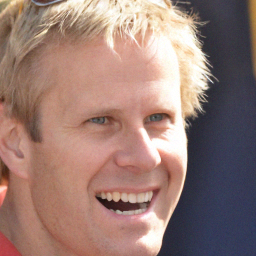} &

        \includegraphics[width=\figwidth]{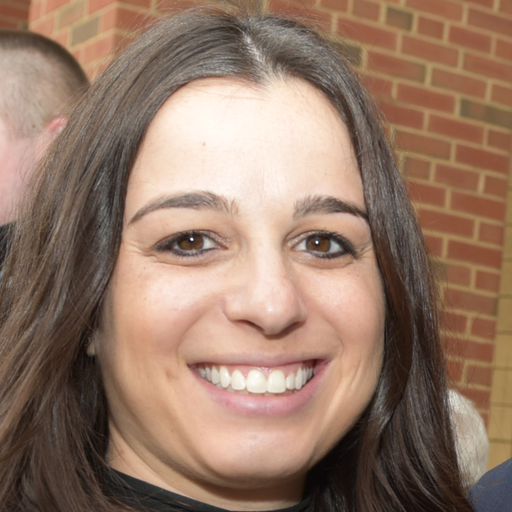} &
        \includegraphics[width=\figwidth]{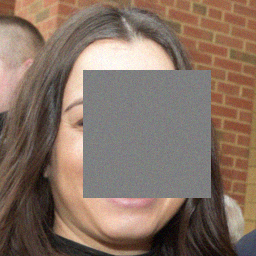} &
        \includegraphics[width=\figwidth]{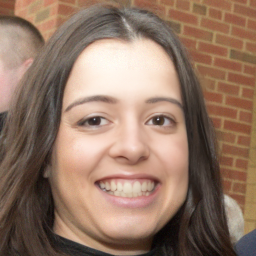} \\

        \includegraphics[width=\figwidth]{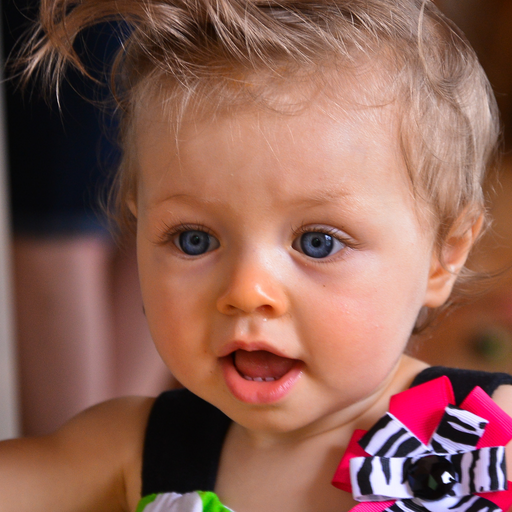} &
        \includegraphics[width=\figwidth]{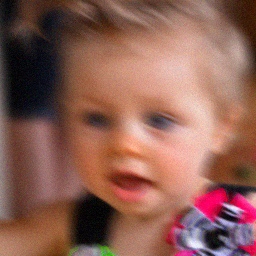} &
        \includegraphics[width=\figwidth]{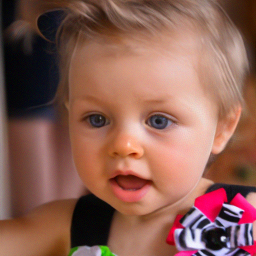} &

        \includegraphics[width=\figwidth]{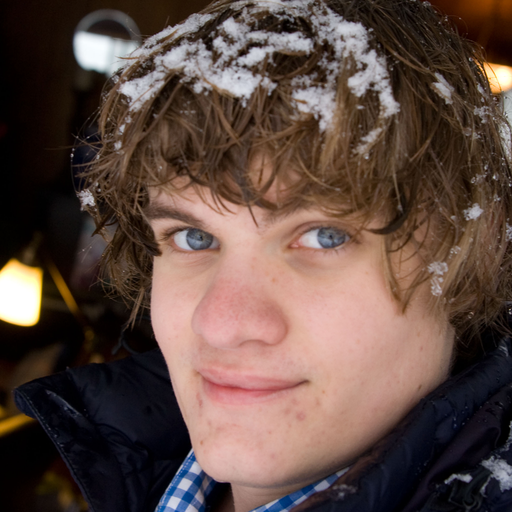} &
        \includegraphics[width=\figwidth]{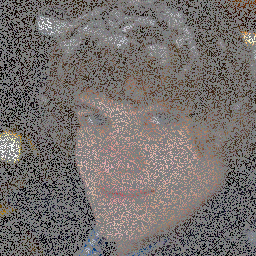} &
        \includegraphics[width=\figwidth]{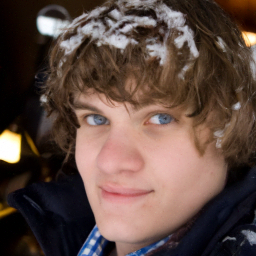} \\

    \end{tabular}%
    \captioning{Qualitative results.}{For each example, we show three images in the following order: ground truth, measurement, and generated sample. On the left half, we show results generated by RePS for nonlinear problems: phase retrieval (top-left) and nonlinear deblurring (bottom-left). On the right half, we show examples for inpainting: box inpaint (top-right) and random inpaint (bottom-right). Visualizations suggest that RePS is able to recover fine details effectively.}

    \label{fig:qualitative}
\end{figure*}
}

\newcommand{\multiruns}{
\begin{figure*}[t]
  \centering
    \newcommand{\figwidth}{0.15\linewidth}%
    \newcommand{\marksize}{2.5}%
    \setlength{\tabcolsep}{0pt}%
    \newcommand{\DATASET}{ffhq}
    \newcommand{\subdir}{\pngdir/multi-runs}%
    \begin{tabular}{@{}ccccccccccc@{}}
         GT & Measurement & Sample-1 & Sample-2 & Sample-3 & Sample-4 \\    

        \includegraphics[width=\figwidth]{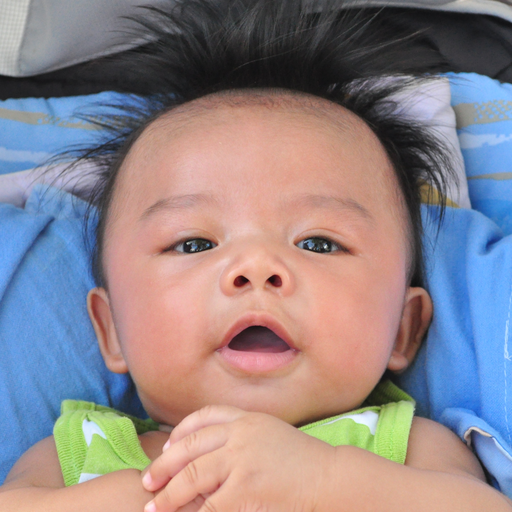} &
        \includegraphics[width=\figwidth]{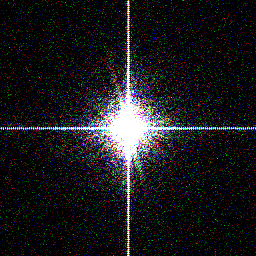} &
        \includegraphics[width=\figwidth]{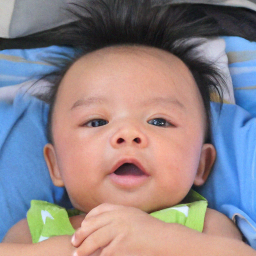} &
        \includegraphics[width=\figwidth]{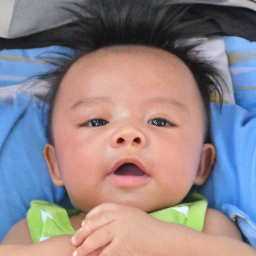} &
        \includegraphics[width=\figwidth]{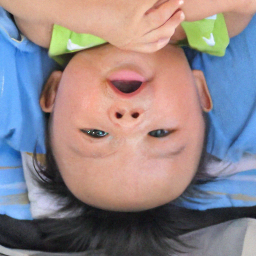} &
        \includegraphics[width=\figwidth]{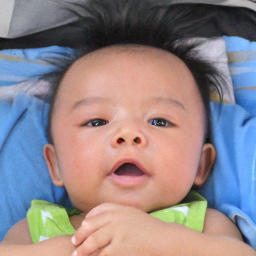} \\

        \includegraphics[width=\figwidth]{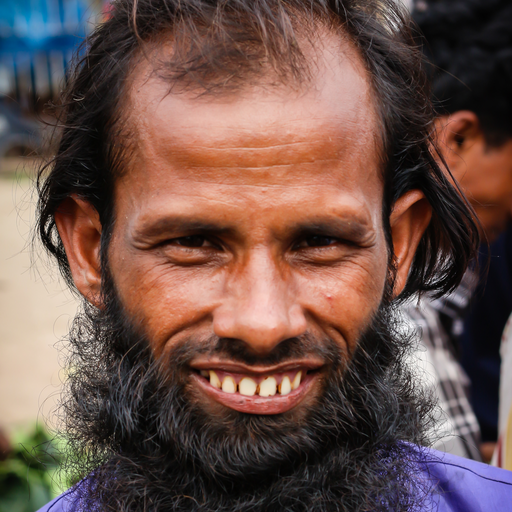} &
        \includegraphics[width=\figwidth]{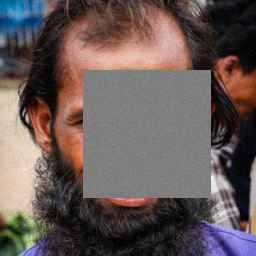} &
        \includegraphics[width=\figwidth]{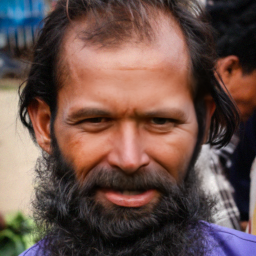} &
        \includegraphics[width=\figwidth]{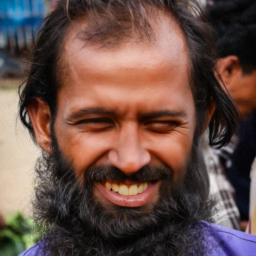} &
        \includegraphics[width=\figwidth]{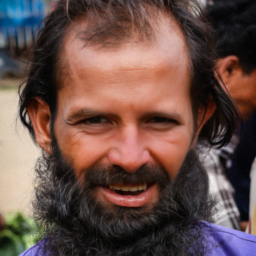} &
        \includegraphics[width=\figwidth]{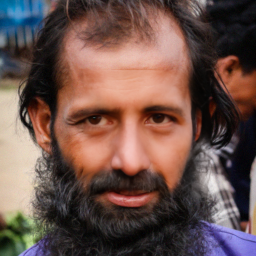} \\
    \end{tabular}%
    \captioning{Multiple samples from the same measurement.}{We show four samples generated from the same measurement for phase retrieval (top row) and inpaint-box (bottom row). The visualizations highlight diverse reconstructions, especially in the masked regions for the inpainting task.}
    \label{fig:multi-runs}
\end{figure*}
}

\newcommand{\phaseRetrievalMultiruns}{
\begin{figure*}[t]
  \centering
    \newcommand{\figwidth}{0.16\linewidth}%
    \newcommand{\marksize}{2.5}%
    \setlength{\tabcolsep}{0pt}%
    \newcommand{\DATASET}{ffhq}
    \newcommand{\subdir}{\pngdir/multi-runs/phase-retrieval}%
    \begin{tabular}{@{}ccccccccccc@{}}
         GT & Measurement & Sample-1 & Sample-2 & Sample-3 & Sample-4 \\    

        \includegraphics[width=\figwidth]{pngs/multi-runs/phase-retrieval/pr1gt} &
        \includegraphics[width=\figwidth]{pngs/multi-runs/phase-retrieval/pr1y} &
        \includegraphics[width=\figwidth]{pngs/multi-runs/phase-retrieval/pr1a} &
        \includegraphics[width=\figwidth]{pngs/multi-runs/phase-retrieval/pr1b} &
        \includegraphics[width=\figwidth]{pngs/multi-runs/phase-retrieval/pr1c} &
        \includegraphics[width=\figwidth]{pngs/multi-runs/phase-retrieval/pr1d} \\

        \includegraphics[width=\figwidth]{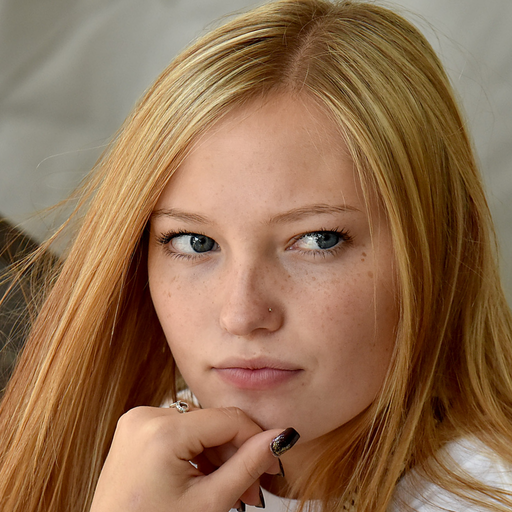} &
        \includegraphics[width=\figwidth]{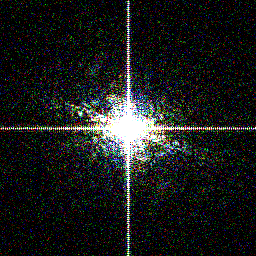} &
        \includegraphics[width=\figwidth]{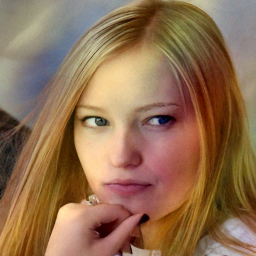} &
        \includegraphics[width=\figwidth]{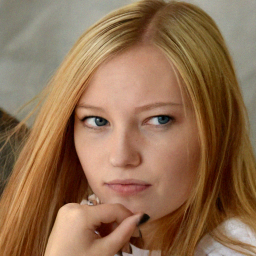} &
        \includegraphics[width=\figwidth]{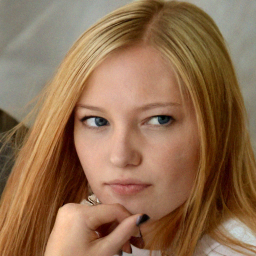} &
        \includegraphics[width=\figwidth]{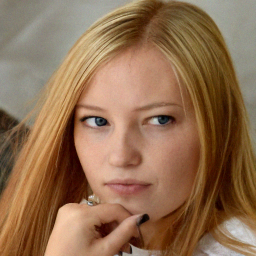} \\

        \includegraphics[width=\figwidth]{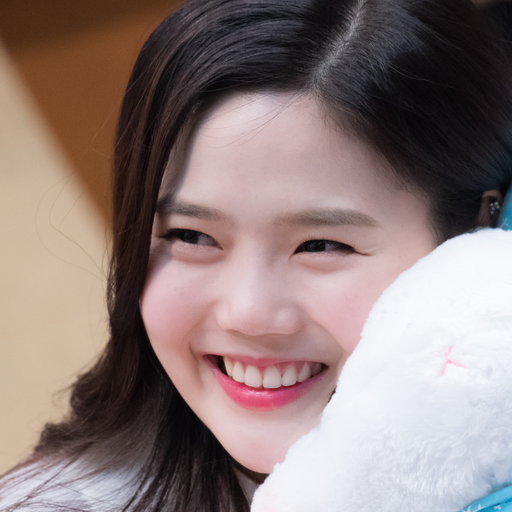} &
        \includegraphics[width=\figwidth]{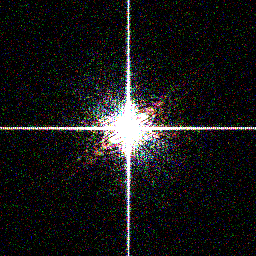} &
        \includegraphics[width=\figwidth]{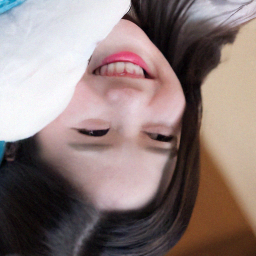} &
        \includegraphics[width=\figwidth]{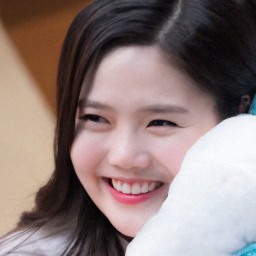} &
        \includegraphics[width=\figwidth]{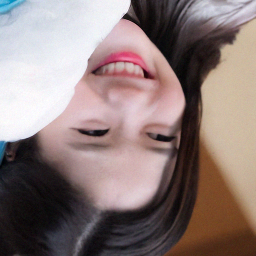} &
        \includegraphics[width=\figwidth]{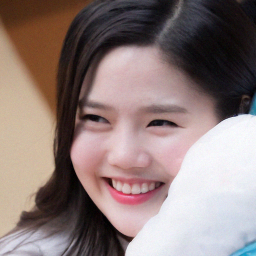} \\

        \includegraphics[width=\figwidth]{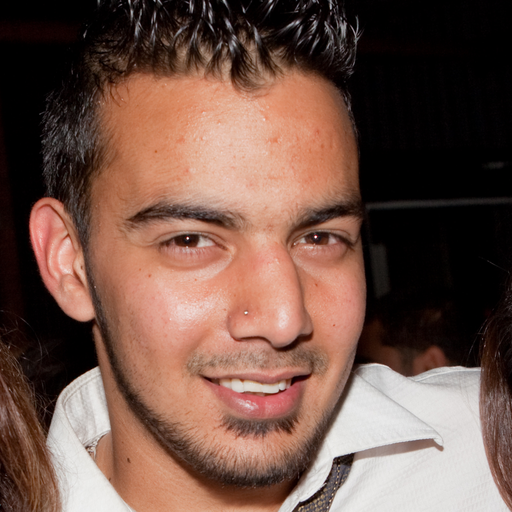} &
        \includegraphics[width=\figwidth]{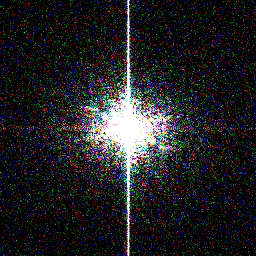} &
        \includegraphics[width=\figwidth]{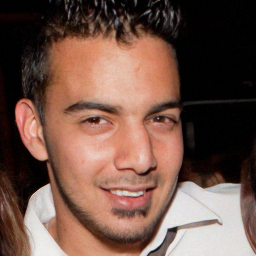} &
        \includegraphics[width=\figwidth]{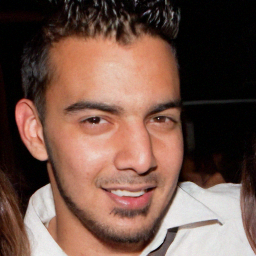} &
        \includegraphics[width=\figwidth]{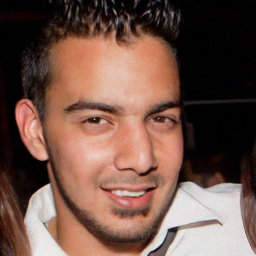} &
        \includegraphics[width=\figwidth]{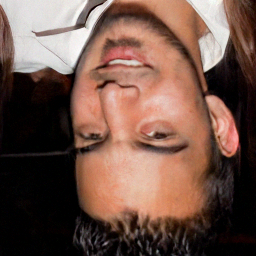} \\
    \end{tabular}%
    \captioning{Multiple generated samples for phase-retrieval.}{Here we show four samples generated from the same measurement for the phase-retrieval problem on FFHQ.}
    \label{fig:phase-retrieval-multiruns}
\end{figure*}
}

\newcommand{\inpaintBoxMultiruns}{
\begin{figure*}[t]
  \centering
    \newcommand{\figwidth}{0.16\linewidth}%
    \newcommand{\marksize}{2.5}%
    \setlength{\tabcolsep}{0pt}%
    \newcommand{\DATASET}{ffhq}
    \newcommand{\subdir}{\pngdir/multi-runs/inpaint-box}%
    \begin{tabular}{@{}ccccccccccc@{}}
         GT & Measurement & Sample-1 & Sample-2 & Sample-3 & Sample-4 \\    

        \includegraphics[width=\figwidth]{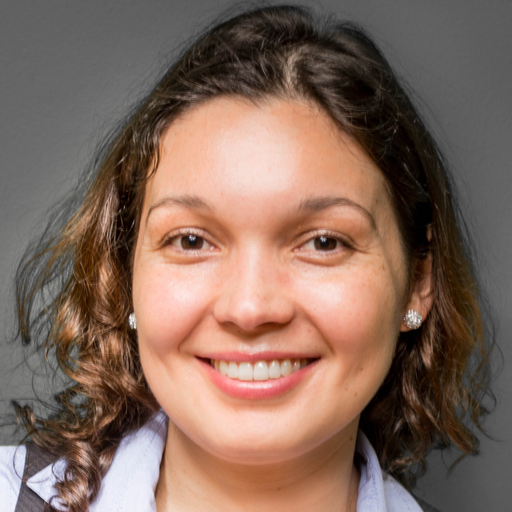} &
        \includegraphics[width=\figwidth]{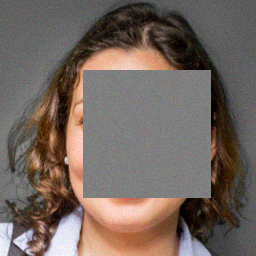} &
        \includegraphics[width=\figwidth]{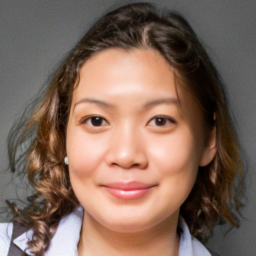} &
        \includegraphics[width=\figwidth]{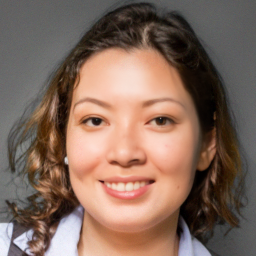} &
        \includegraphics[width=\figwidth]{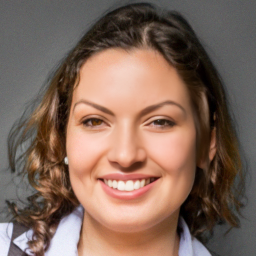} &
        \includegraphics[width=\figwidth]{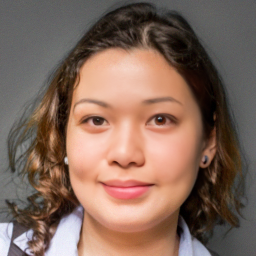} \\

        \includegraphics[width=\figwidth]{pngs/multi-runs/inpaint-box/ip2gt} &
        \includegraphics[width=\figwidth]{pngs/multi-runs/inpaint-box/ip2y} &
        \includegraphics[width=\figwidth]{pngs/multi-runs/inpaint-box/ip2a} &
        \includegraphics[width=\figwidth]{pngs/multi-runs/inpaint-box/ip2b} &
        \includegraphics[width=\figwidth]{pngs/multi-runs/inpaint-box/ip2c} &
        \includegraphics[width=\figwidth]{pngs/multi-runs/inpaint-box/ip2d} \\

        \includegraphics[width=\figwidth]{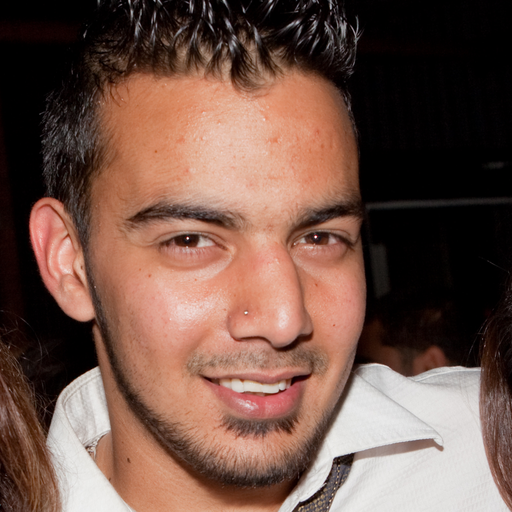} &
        \includegraphics[width=\figwidth]{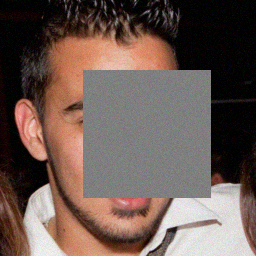} &
        \includegraphics[width=\figwidth]{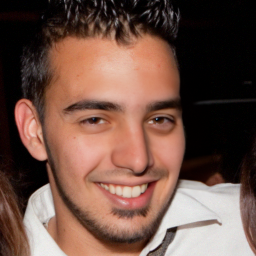} &
        \includegraphics[width=\figwidth]{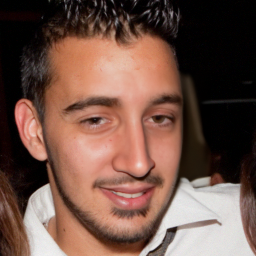} &
        \includegraphics[width=\figwidth]{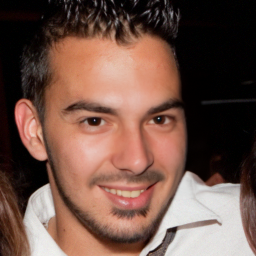} &
        \includegraphics[width=\figwidth]{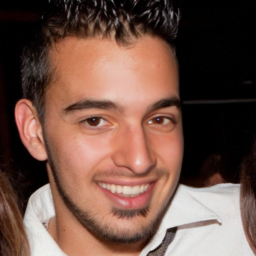} \\

        \includegraphics[width=\figwidth]{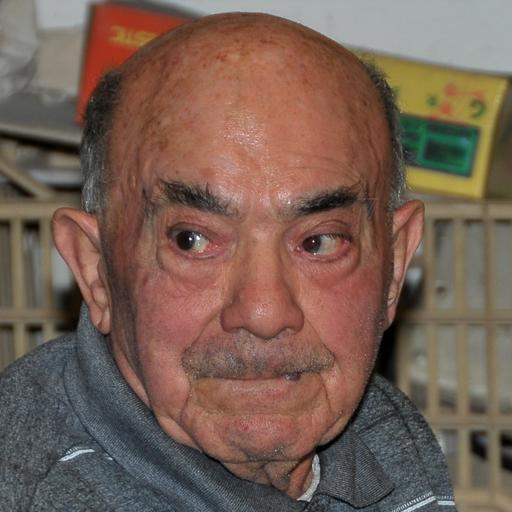} &
        \includegraphics[width=\figwidth]{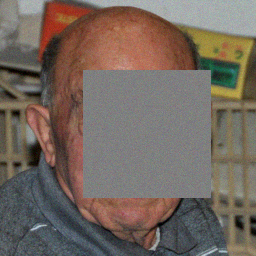} &
        \includegraphics[width=\figwidth]{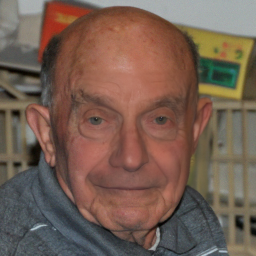} &
        \includegraphics[width=\figwidth]{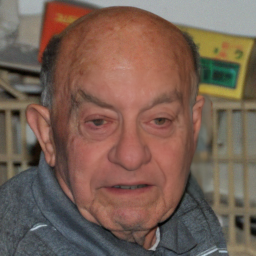} &
        \includegraphics[width=\figwidth]{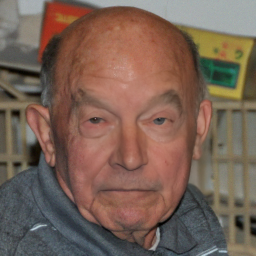} &
        \includegraphics[width=\figwidth]{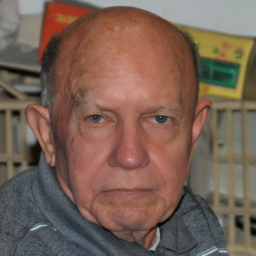} \\
    \end{tabular}%
    \captioning{Multiple generated samples for box inpainting.}{Here we show four samples generated from the same measurement for the box inpainting problem on FFHQ.}
    \label{fig:inpaint-box-multiruns}
\end{figure*}
}

\newcommand{\FFHQFiguresOne}{
\begin{figure*}[t]
  \centering
    \newcommand{\figwidth}{0.16\linewidth}%
    \newcommand{\marksize}{2.5}%
    \setlength{\tabcolsep}{0pt}%
    \newcommand{\DATASET}{ffhq}
    \newcommand{\subdir}{\pngdir/\DATASET}%
    \begin{tabular}{@{}ccccccccccc@{}}
         \multicolumn{3}{c}{Inpaint (Box)} & \multicolumn{3}{c}{Inpaint (Random)} \\    

        \includegraphics[width=\figwidth]{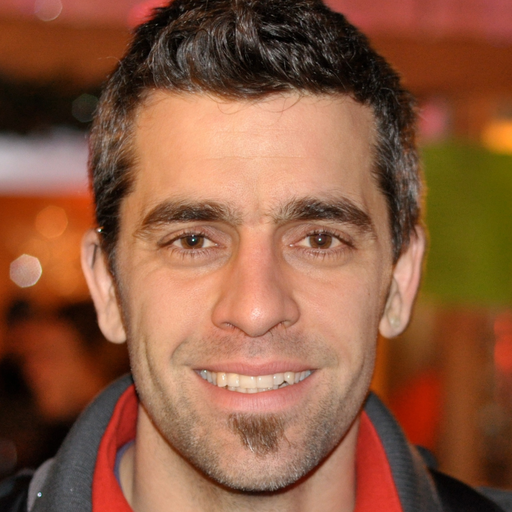} &
        \includegraphics[width=\figwidth]{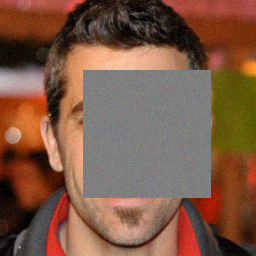} &
        \includegraphics[width=\figwidth]{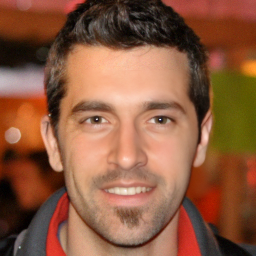} &

        \includegraphics[width=\figwidth]{pngs/\DATASET/inpaint-random/1gt} &
        \includegraphics[width=\figwidth]{pngs/\DATASET/inpaint-random/1y} &
        \includegraphics[width=\figwidth]{pngs/\DATASET/inpaint-random/1gen} \\

        \includegraphics[width=\figwidth]{pngs/\DATASET/inpaint-box/2gt} &
        \includegraphics[width=\figwidth]{pngs/\DATASET/inpaint-box/2y} &
        \includegraphics[width=\figwidth]{pngs/\DATASET/inpaint-box/2gen} &

        \includegraphics[width=\figwidth]{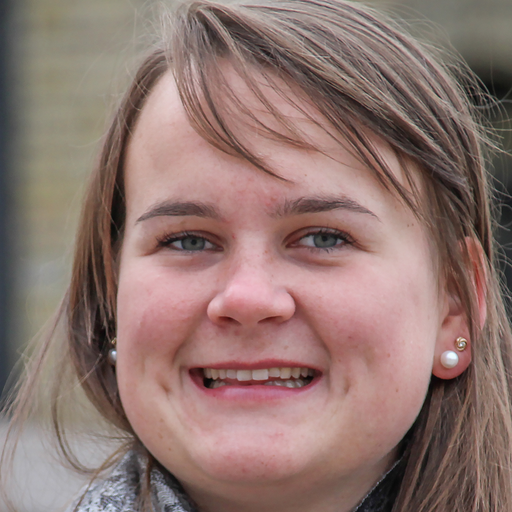} &
        \includegraphics[width=\figwidth]{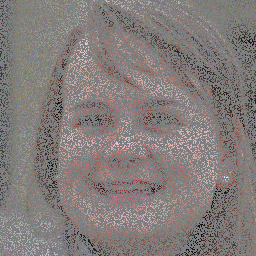} &
        \includegraphics[width=\figwidth]{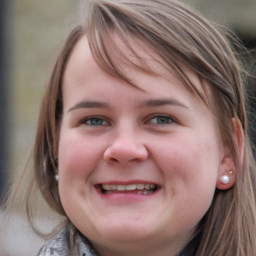} \\

        \includegraphics[width=\figwidth]{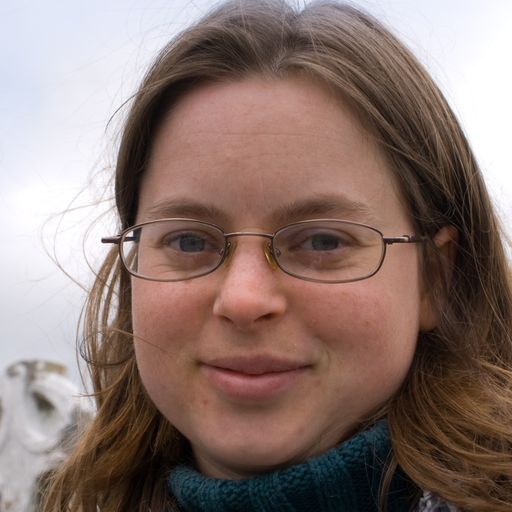} &
        \includegraphics[width=\figwidth]{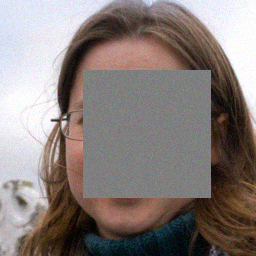} &
        \includegraphics[width=\figwidth]{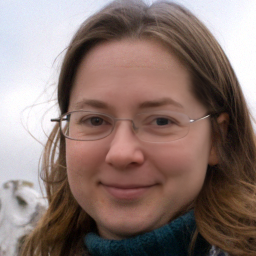} &

        \includegraphics[width=\figwidth]{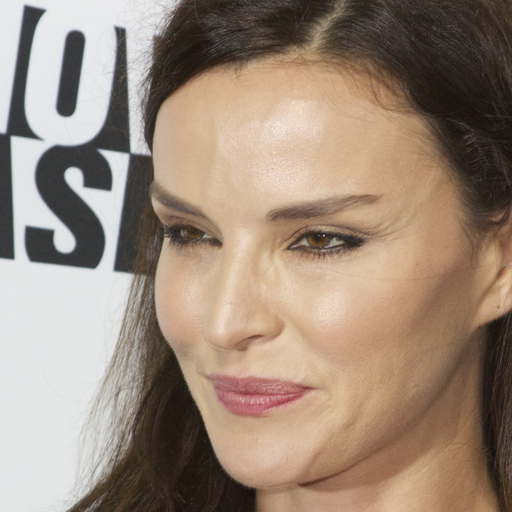} &
        \includegraphics[width=\figwidth]{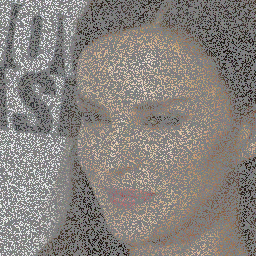} &
        \includegraphics[width=\figwidth]{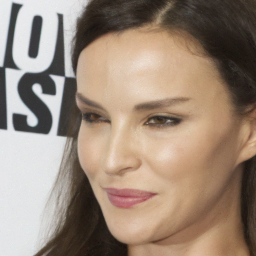} \\
    \end{tabular}%
    \captioning{Qualitative results for Inpaint-Box (Left) and Inpaint-Random (right) on FFHQ.}{Three images are displayed for each example in the order: the ground truth, the measurement, a sample generated by RePS.}
    \label{fig:qualitative-ffhq1}
\end{figure*}
}

\newcommand{\FFHQFiguresTwo}{
\begin{figure*}[t]
  \centering
    \newcommand{\figwidth}{0.16\linewidth}%
    \newcommand{\marksize}{2.5}%
    \setlength{\tabcolsep}{0pt}%
    \newcommand{\DATASET}{ffhq}
    \newcommand{\subdir}{\pngdir/\DATASET}%
    \begin{tabular}{@{}ccccccccccc@{}}
         \multicolumn{3}{c}{Gaussian Deblur} & \multicolumn{3}{c}{Motion Deblur} \\

        \includegraphics[width=\figwidth]{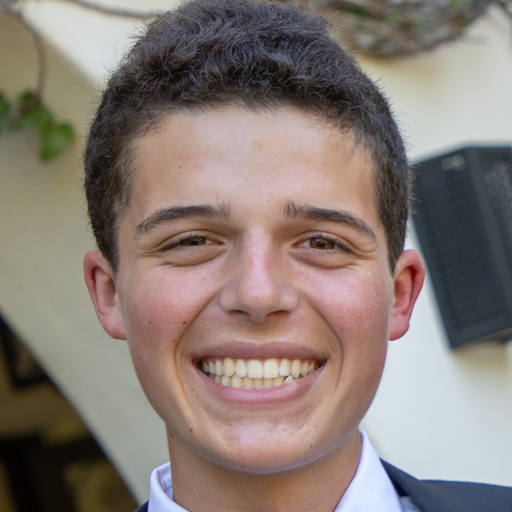} &
        \includegraphics[width=\figwidth]{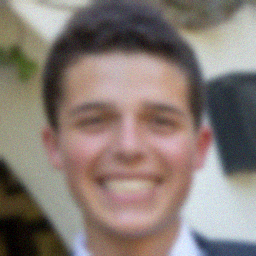} &
        \includegraphics[width=\figwidth]{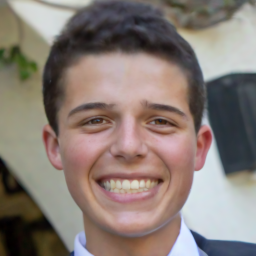} &

        \includegraphics[width=\figwidth]{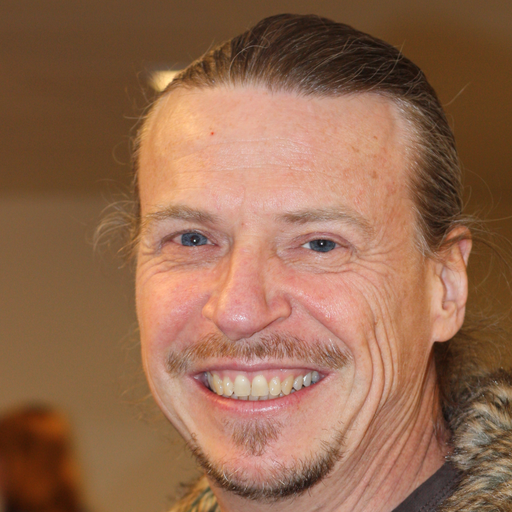} &
        \includegraphics[width=\figwidth]{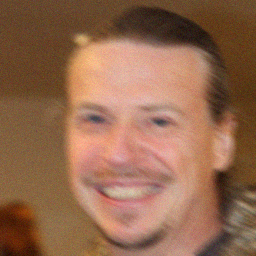} &
        \includegraphics[width=\figwidth]{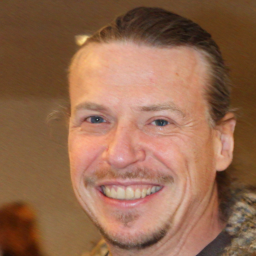} \\

        \includegraphics[width=\figwidth]{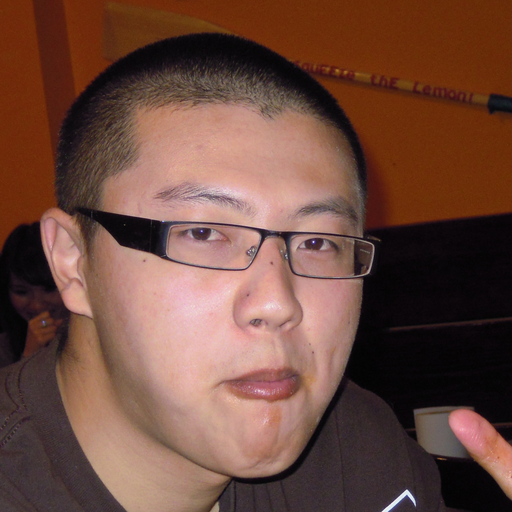} &
        \includegraphics[width=\figwidth]{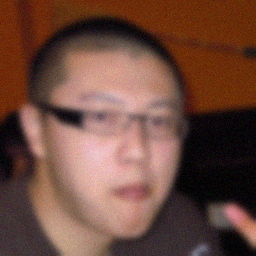} &
        \includegraphics[width=\figwidth]{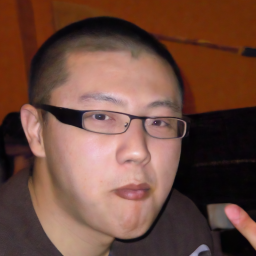} &

        \includegraphics[width=\figwidth]{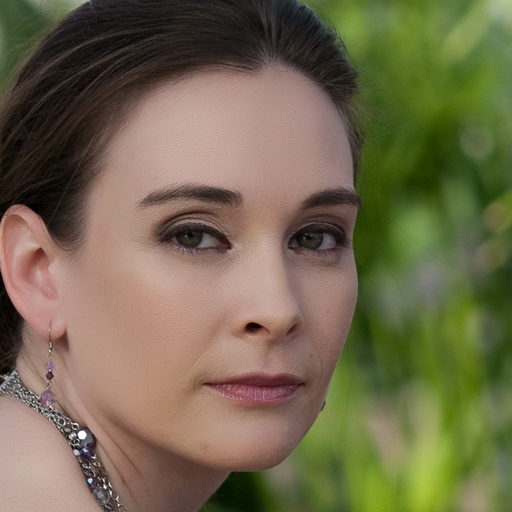} &
        \includegraphics[width=\figwidth]{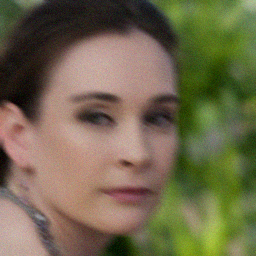} &
        \includegraphics[width=\figwidth]{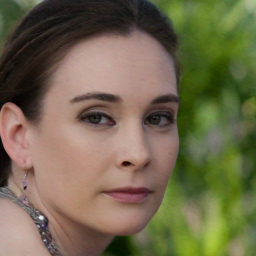} \\

        \includegraphics[width=\figwidth]{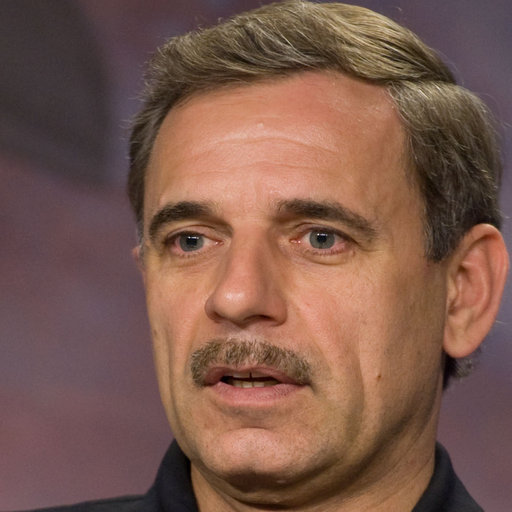} &
        \includegraphics[width=\figwidth]{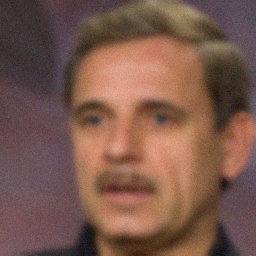} &
        \includegraphics[width=\figwidth]{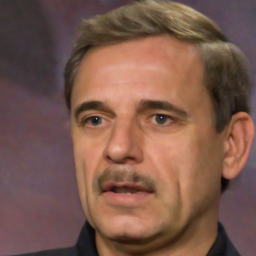} &

        \includegraphics[width=\figwidth]{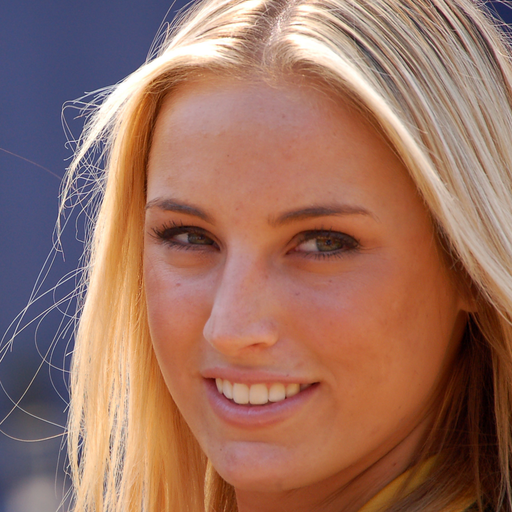} &
        \includegraphics[width=\figwidth]{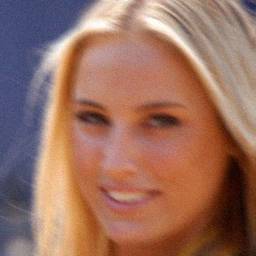} &
        \includegraphics[width=\figwidth]{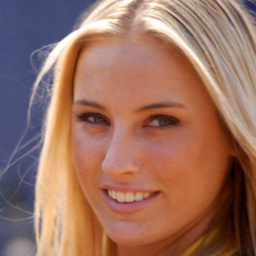} \\
    \end{tabular}%
    \captioning{Qualitative results for Gaussian Deblur (left) and Motion Deblur (right) on FFHQ.}{Three images are displayed for each example in the order: the ground truth, the measurement, a sample generated by RePS.}
    \label{fig:qualitative-ffhq2}
\end{figure*}
}
\newcommand{\FFHQFiguresThree}{
\begin{figure*}[t]
  \centering
    \newcommand{\figwidth}{0.16\linewidth}%
    \newcommand{\marksize}{2.5}%
    \setlength{\tabcolsep}{0pt}%
    \newcommand{\DATASET}{ffhq}
    \newcommand{\subdir}{\pngdir/\DATASET}%
    \begin{tabular}{@{}ccccccccccc@{}}
         \multicolumn{3}{c}{Super Resolution} & \multicolumn{3}{c}{HDR} \\    

        \includegraphics[width=\figwidth]{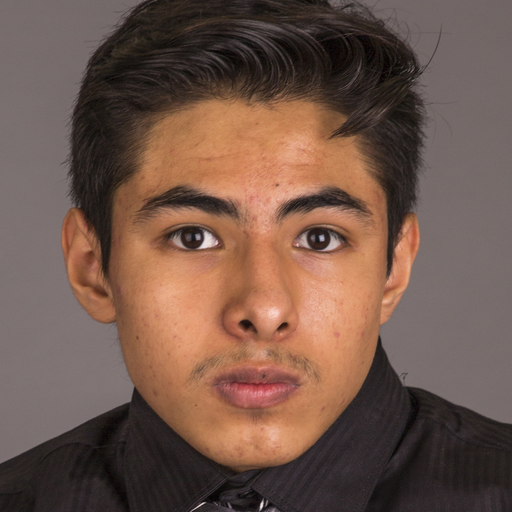} &
        \includegraphics[width=\figwidth]{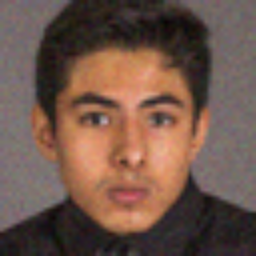} &
        \includegraphics[width=\figwidth]{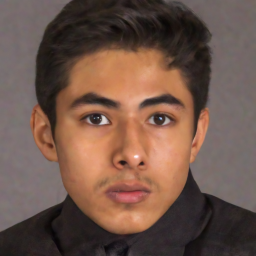} &

        \includegraphics[width=\figwidth]{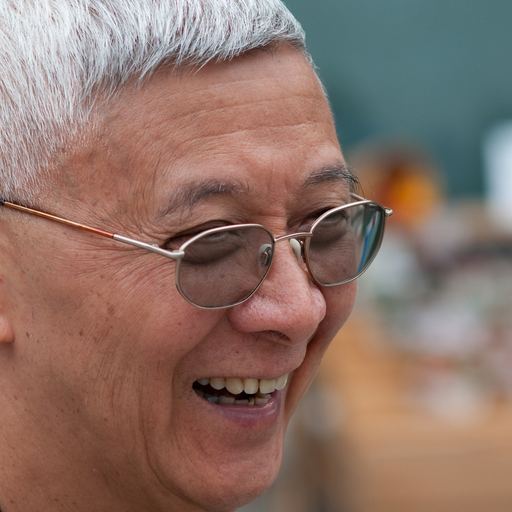} &
        \includegraphics[width=\figwidth]{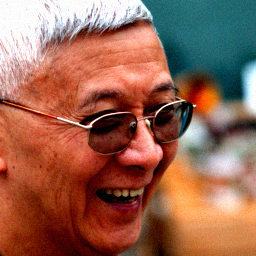} &
        \includegraphics[width=\figwidth]{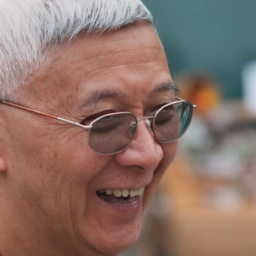} \\

        \includegraphics[width=\figwidth]{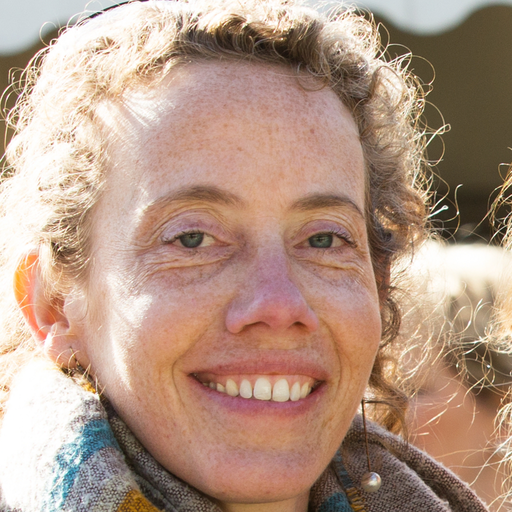} &
        \includegraphics[width=\figwidth]{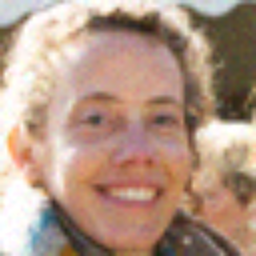} &
        \includegraphics[width=\figwidth]{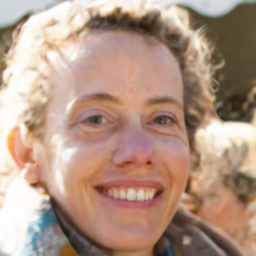} &

        \includegraphics[width=\figwidth]{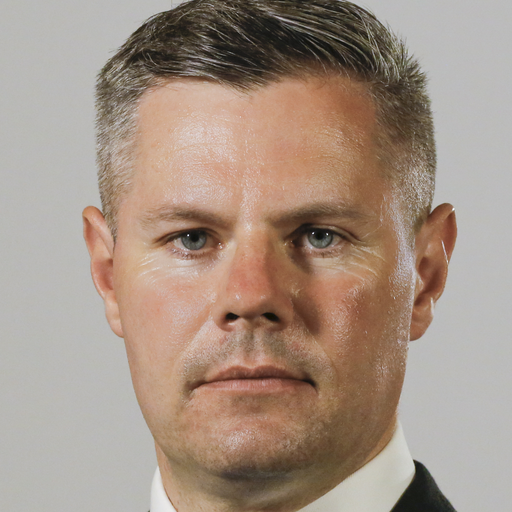} &
        \includegraphics[width=\figwidth]{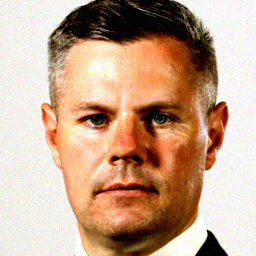} &
        \includegraphics[width=\figwidth]{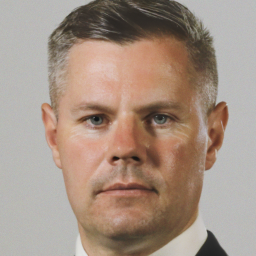} \\

        \includegraphics[width=\figwidth]{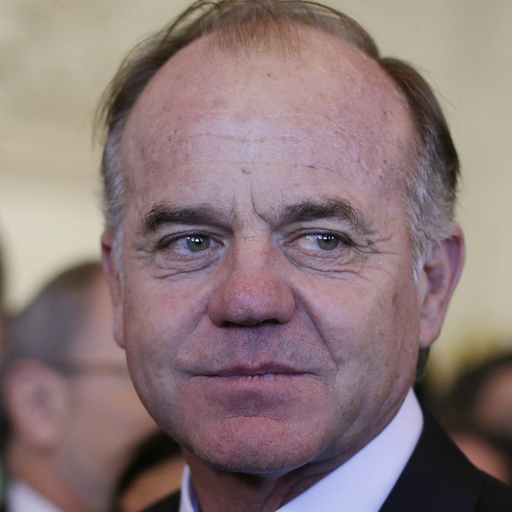} &
        \includegraphics[width=\figwidth]{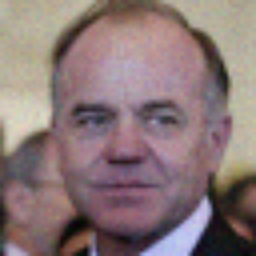} &
        \includegraphics[width=\figwidth]{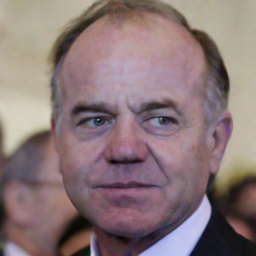} &

        \includegraphics[width=\figwidth]{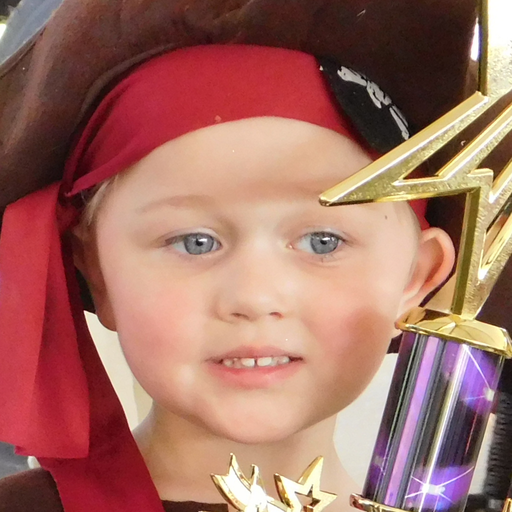} &
        \includegraphics[width=\figwidth]{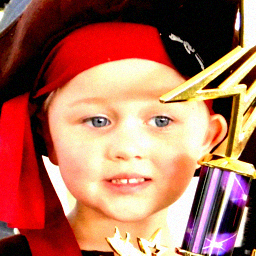} &
        \includegraphics[width=\figwidth]{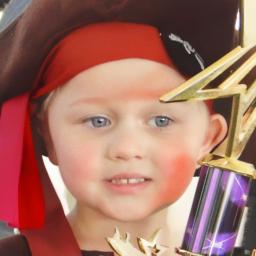} \\
    \end{tabular}%
    \captioning{Qualitative results for Super Resolution (left) and HDR (right) on FFHQ.}{Three images are displayed for each example in the order: the ground truth, the measurement, a sample generated by RePS.}
    \label{fig:qualitative-ffhq3}
\end{figure*}
}

\newcommand{\FFHQFiguresFour}{
\begin{figure*}[t]
  \centering
    \newcommand{\figwidth}{0.16\linewidth}%
    \newcommand{\marksize}{2.5}%
    \setlength{\tabcolsep}{0pt}%
    \newcommand{\DATASET}{ffhq}
    \newcommand{\subdir}{\pngdir/\DATASET}%
    \begin{tabular}{@{}ccccccccccc@{}}
         \multicolumn{3}{c}{Phase Retrieval} & \multicolumn{3}{c}{Nonlinear Deblur} \\    

        \includegraphics[width=\figwidth]{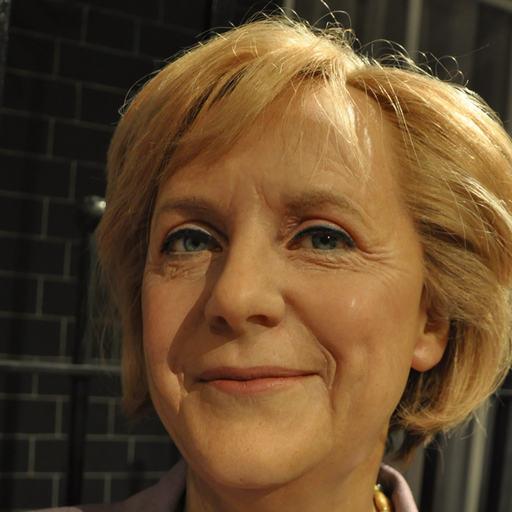} &
        \includegraphics[width=\figwidth]{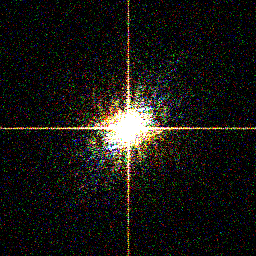} &
        \includegraphics[width=\figwidth]{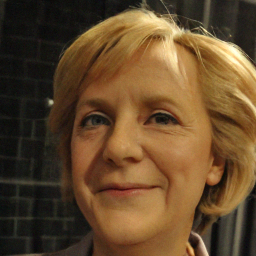} &

        \includegraphics[width=\figwidth]{pngs/\DATASET/nonlinear-deblur/1gt} &
        \includegraphics[width=\figwidth]{pngs/\DATASET/nonlinear-deblur/1y} &
        \includegraphics[width=\figwidth]{pngs/\DATASET/nonlinear-deblur/1gen} \\

        \includegraphics[width=\figwidth]{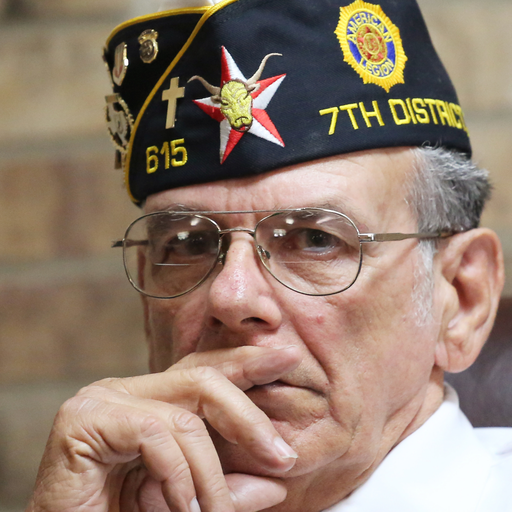} &
        \includegraphics[width=\figwidth]{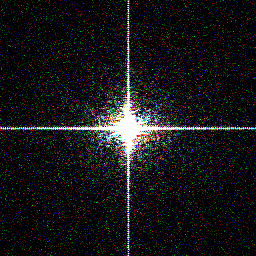} &
        \includegraphics[width=\figwidth]{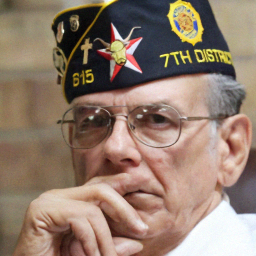} &

        \includegraphics[width=\figwidth]{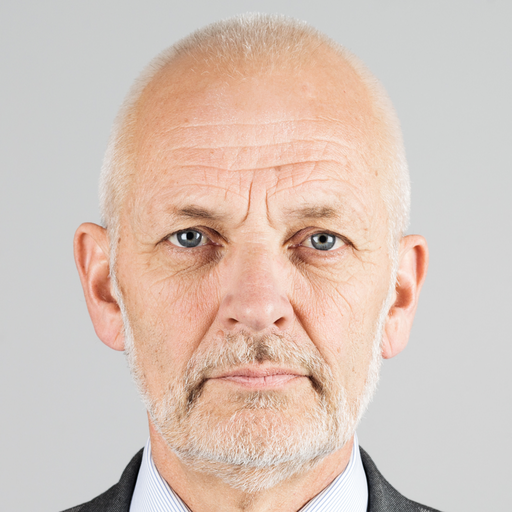} &
        \includegraphics[width=\figwidth]{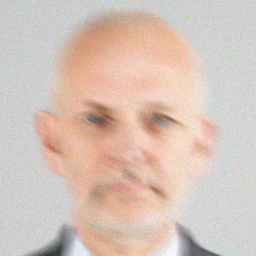} &
        \includegraphics[width=\figwidth]{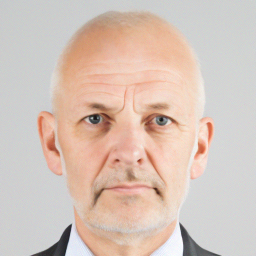} \\

        \includegraphics[width=\figwidth]{pngs/\DATASET/phase-retrieval/3gt} &
        \includegraphics[width=\figwidth]{pngs/\DATASET/phase-retrieval/3y} &
        \includegraphics[width=\figwidth]{pngs/\DATASET/phase-retrieval/3gen} &

        \includegraphics[width=\figwidth]{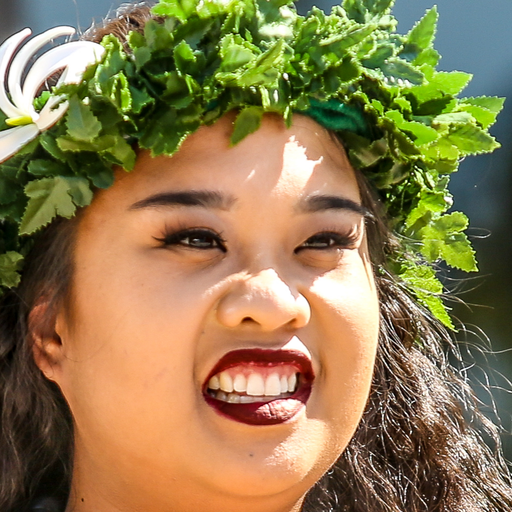} &
        \includegraphics[width=\figwidth]{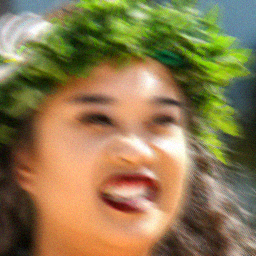} &
        \includegraphics[width=\figwidth]{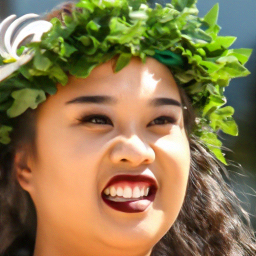} \\
    \end{tabular}%
    \captioning{Qualitative results for Phase Retrieval (left) and Nonlinear Deblur (right) on FFHQ.}{Three images are displayed for each example in the order: the ground truth, the measurement, a sample generated by RePS.}
    \label{fig:qualitative-ffhq4}
\end{figure*}
}

\newcommand{\repsAlgo}{%
\begin{algorithm}[t]
  \caption{Sampling with conditioned ODE + restart}\label{alg:reps}
  \begin{algorithmic}[1]
    \Require $s_{\theta}$, $\sigma_{\max}$, $\sigma_{0}$, $\sigma_{\text{res}}$, $\sigma_{\min}$, $N_{\text{res}}$, $\rho_{\text{res}}$, $N_{\text{ode}}$, $\rho_{\text{ode}}$
    
    \State Restart schedule: $\sigma(r)$ = schedule($\sigma_{\text{res}}$, $\sigma_{\min}$, $N_{\text{res}}$, $\rho_{\text{res}}$) 
    \State Prepend $\sigma_{\max}$ to restart schedule: $\sigma(r_0) = \sigma_{\max}$
    \State Sample $\mathbf{x}_{r_0} \sim \mathcal{N}(0, \sigma^2_{\max} I)$
    \For{$r \in$ Restart schedule}
      \State ODE schedule: $\sigma(t)$ = schedule($\sigma_{r}$, $\sigma_{0}$, $N_{\text{ode}}$, $\rho_{\text{ode}}$) 
      \State $\rv{x}_t = \rv{x}_{r}$
      \For{$t \in$ ODE schedule}
        \State $\hat{\rv{x}}_{0} = \mathbb{E}[\rv{x}_0 \mid \rv{x}_t]$ using $s_{\theta}$
        \State MAP estimate by solving \eqn{map-optimization} 
        \State $\rv{x}_{\,t-1} = \rv{x}_{\,t} + \frac{\sigma(t) - \sigma(t-1)}{\sigma(t)}\;(\rv{x}_{\rm MAP} - \rv{x}_{\,t})$
        
      \EndFor
      \State Restart: $\rv{x}_{r+1} \sim \mathcal{N}(\hat{\rv{x}}_{0\mid y}, \sigma^2(r+1)\, I)$
    \EndFor
  \end{algorithmic}
\end{algorithm}%
}

\newcommand{\resultsSummary}{
\begin{table*}[t]
\centering
\captioning{Quantitative results for FFHQ and Imagenet on a set of linear and nonlinear problems.}{
}
\label{tab:main_results}
\resizebox{0.9\textwidth}{!}{%
\begin{tabular}{@{}llcccccccc@{}}
\toprule
\multicolumn{1}{c}{Task} &  \multicolumn{1}{c}{Method} & \multicolumn{4}{c}{FFHQ} & \multicolumn{4}{c}{ImageNet} \\
\cmidrule(lr){3-6} \cmidrule(lr){7-10} 
 & & PSNR ($\uparrow$) & SSIM ($\uparrow$) & LPIPS ($\downarrow$) & FID ($\downarrow$) & PSNR ($\uparrow$) & SSIM ($\uparrow$) & LPIPS ($\downarrow$) & FID ($\downarrow$) \\ 
 
\midrule
\multirow{2}{*}{Super Resolution 4$\times$} 
& RePS (ours) & \textbf{29.95} & \textbf{0.845} & \textbf{0.155} & \textbf{48.37}  & \textbf{26.12} & \textbf{0.708} & \textbf{0.259} & 114.19 \\

& \citet{li_efficient_2025} & 29.21 & 0.805 & 0.256 & 94.42  & 25.68 & 0.682 & 0.307 & 135.39  \\

& DAPS (code) & \underline{29.55} & 0.793 & 0.186 & 52.95  & \underline{25.70} & 0.654 & \underline{0.285} & 106.50  \\

& DPS & 25.86 & 0.753 & 0.269 & 81.07 & 21.13 & 0.489 & 0.361 & 106.32 \\
& DDRM & 26.58 & 0.782 & 0.282 & 79.25 & 22.62 & 0.521 & 0.324 & 103.85 \\
& DDNM & 28.03 & 0.795 & 0.197 & 64.62 & 23.96 & 0.604 & 0.475 & \underline{98.62} \\
& DCDP & 28.66 & 0.807 & \underline{0.178} & 53.81 & - & - & - & - \\
& FPS-SMC & 28.42 & \underline{0.813} & 0.204 & \underline{49.25} & 24.82 &  \underline{0.703} & 0.313 &  \textbf{97.51} \\
& DiffPIR & 26.64 & - & 0.260 & 65.77 & 23.18 & - & 0.371 & 106.32 \\

\midrule
\multirow{2}{*}{Inpaint (Box)} 
& RePS (ours) & 24.32 & \textbf{0.844} & \textbf{0.122} & \textbf{41.20} & 20.71 & \textbf{0.787} & \textbf{0.186} & \underline{109.21} \\

& \citet{li_efficient_2025} & 24.75 & 0.817 & 0.167 & 57.67  & 21.39 & \underline{0.779} & 0.217 & 128.54 \\

& DAPS (code) & \textbf{24.88} & 0.755 & 0.174 & 49.83  & 21.40 & 0.721 & 0.228 & 113.96 \\

& DPS & 22.51 & 0.792 & 0.209 & 61.27 & 18.94 & 0.722 & 0.257 & 126.52 \\
& DDRM & 22.26 & 0.801 & 0.207 & 78.62 & 18.63 & 0.733 & 0.254 & 116.37 \\
& DDNM & 24.47 & \underline{0.837} & 0.235 & 46.59 & \underline{21.64} & 0.748 & 0.319 & \textbf{103.97} \\
& DCDP & 23.89 & 0.760 & 0.163 & \underline{45.23} & - & - & - & - \\
& FPS-SMC & \underline{24.86} & 0.823 & \underline{0.146} & 48.34 & \textbf{22.16} & 0.726 & \underline{0.208} & 111.58 \\

\midrule
\multirow{1}{*}{Inpaint (Random)}  
& RePS (ours) & \underline{32.49} & \underline{0.899} & \underline{0.090} & \underline{24.18} & 28.97 & 0.829 & \underline{0.115} & \textbf{27.52} \\

& \citet{li_efficient_2025} & \textbf{34.11} &  \textbf{0.926} &  \textbf{0.071} & \textbf{20.04}  & \underline{29.67} & \textbf{0.861} & \textbf{0.114} & \underline{29.70} \\

&  DAPS (code) & 30.94 & 0.807 & 0.155 & 34.84  & 27.72 & 0.742 & 0.177 & 37.18 \\
& DPS & 25.46 & 0.823 & 0.203 & 69.20 & 23.52 & 0.745 & 0.297 & 87.53 \\
& DDNM & 29.91 & 0.817 & 0.121 & 44.37 & \textbf{31.16} & \underline{0.841} & 0.191 & 63.84 \\
& DCDP & 30.69 & 0.842 & 0.142 & 52.51 & - & - & - & - \\
& FPS-SMC & 28.21 & 0.823 & 0.261 & 61.23 & 24.52 & 0.701 & 0.316 & 79.12 \\

\midrule
\multirow{1}{*}{Gaussian deblurring} 
& RePS (ours) & \textbf{29.92} & \textbf{0.841} & \textbf{0.146} & \textbf{49.68} & \underline{26.21} & \textbf{0.703} & \textbf{0.255} & 116.84 \\

& \citet{li_efficient_2025} & \underline{29.89} & \underline{0.831} & 0.217 & 86.54 & 25.75 & \underline{0.681} & 0.296 & 143.88 \\

& DAPS (code) & 29.75 & 0.794 & \underline{0.177} & \underline{53.01} & 26.00 & 0.668 & \underline{0.260} & 102.46 \\
& DPS & 25.87 & 0.764 & 0.219 & 79.75 & 20.31 & 0.598 & 0.397 & 116.42 \\
& DDRM & 24.93 & 0.732 & 0.239 & 92.43 & 21.26 & 0.564 & 0.443 & 146.89 \\
& DDNM & 28.20 & 0.804 & 0.216 & 57.83 & \textbf{28.06} & \textbf{0.703} & 0.278 & \textbf{81.43} \\
& DCDP & 27.50 & 0.699 & 0.304 & 86.43 & - & - & - & - \\
& FPS-SMC & 26.54 & 0.773 & 0.253 & 67.45 & 23.91 & 0.601 & 0.387 & \underline{91.72} \\
& DiffPIR & 27.36 & - & 0.236 & 59.65 & 22.80 & - & 0.355 & 93.36 \\

\midrule
\multirow{1}{*}{Motion deblurring}
& RePS (ours) & \textbf{32.27} & \textbf{0.875} & \textbf{0.115} & \textbf{34.10} & \textbf{28.95} & \textbf{0.801} & \textbf{0.169} & \underline{53.15} \\

& \citet{li_efficient_2025} & 29.90 & 0.766 & 0.207 & 62.43 & 27.70 & 0.733 & 0.238 & 79.27 \\

& DAPS (code) & \underline{31.80} & \underline{0.843} & \underline{0.135} & \underline{40.82}  & \underline{28.87} & \underline{0.781} & \underline{0.171} & \textbf{50.35} \\
& DPS & 24.52 & 0.801 & 0.246 & 65.23 & 18.96 & 0.629 & 0.423 & 137.81 \\
& DCDP & 25.08 & 0.512 & 0.364 & 125.13 & - & - & - & - \\
& FPS-SMC & 27.39 & 0.826 & 0.227 & 48.32 & 24.52 & 0.647 & 0.326 & 87.43 \\
& DiffPIR & 26.57 & - & 0.255 & 65.78 & 24.01 & - & 0.366 & 94.63 \\

\midrule
\multirow{1}{*}{Phase Retrieval} 

& RePS (ours)  & \underline{30.41}\tiny{$\pm$4.45} & \textbf{0.824}\tiny{$\pm$0.119} & \textbf{0.152}\tiny{$\pm$0.107} &  \underline{41.81} & \underline{20.12}\tiny{$\pm$7.18} & \underline{0.449}\tiny{$\pm$0.260} & \underline{0.419}\tiny{$\pm$0.184} & 220.93 \\
& DAPS (code) & \textbf{30.72}\tiny{$\pm$3.15} & \underline{0.809}\tiny{$\pm$0.081} & \underline{0.157}\tiny{$\pm$0.073} & \textbf{37.88} & \textbf{22.32}\tiny{$\pm$6.51} & \textbf{0.514}\tiny{$\pm$0.219} & \textbf{0.343}\tiny{$\pm$0.155} & \textbf{134.55} \\
& DPS & 17.64 \tiny{$\pm$2.97} & 0.441\tiny{$\pm$0.129} & 0.410\tiny{$\pm$0.090} & 104.52 & 16.81\tiny{$\pm$3.61} & 0.427\tiny{$\pm$0.143} & 0.447\tiny{$\pm$0.099} & \underline{197.54} \\
& RED-diff & 15.60 \tiny{$\pm$4.48} & 0.398 \tiny{$\pm$0.195} & 0.596 \tiny{$\pm$0.092} & 167.43 & 14.98\tiny{$\pm$3.75} & 0.386\tiny{$\pm$0.057} & 0.536\tiny{$\pm$0.129} & 212.24 \\
& DCDP & 28.65 \tiny{$\pm$8.09} & 0.781 \tiny{$\pm$0.217} & 0.203 \tiny{$\pm$0.196} & 68.13 & - & - & - & - \\

\midrule
\multirow{1}{*}{Nonlinear deblur} 
& RePS (ours)  & \underline{29.02}\tiny{$\pm$1.76} & \textbf{0.797}\tiny{$\pm$0.031} & \underline{0.165}\tiny{$\pm$0.030} & 51.97 & \underline{27.58}\tiny{$\pm$3.28} & \underline{0.745}\tiny{$\pm$0.082} & \textbf{0.191}\tiny{$\pm$0.056} & 72.66 \\
& DAPS (code) & 28.79\tiny{$\pm$1.54} & 0.781\tiny{$\pm$0.033} & 0.177\tiny{$\pm$0.030} & \underline{51.08} & 27.47\tiny{$\pm$3.21} & 0.737\tiny{$\pm$0.083} & \underline{0.198}\tiny{$\pm$0.052} & \underline{67.62} \\
& DPS & 23.39\tiny{$\pm$2.01} & 0.623\tiny{$\pm$0.082} & 0.278\tiny{$\pm$0.060} & 91.31 & 22.49\tiny{$\pm$3.20} & 0.591\tiny{$\pm$0.101} & 0.306\tiny{$\pm$0.080} & 101.41 \\
& RED-diff & \textbf{30.86}\tiny{$\pm$0.51} & \underline{0.795}\tiny{$\pm$0.028} & \textbf{0.160}\tiny{$\pm$0.034} & \textbf{43.84} & \textbf{30.07}\tiny{$\pm$1.41} & \textbf{0.754}\tiny{$\pm$0.023} & 0.211\tiny{$\pm$0.083} & \textbf{51.22} \\
& DCDP & 27.92\tiny{$\pm$2.64} & 0.779\tiny{$\pm$0.067} & 0.183\tiny{$\pm$0.051} & 51.96 & - & - & - & - \\

\midrule
\multirow{1}{*}{High dynamic range} 
& RePS (ours)  & \textbf{27.96}\tiny{$\pm$3.54} & \textbf{0.872}\tiny{$\pm$0.080} & \textbf{0.145}\tiny{$\pm$0.071} & \textbf{36.01} & \underline{26.37}\tiny{$\pm$4.05} & \textbf{0.843}\tiny{$\pm$0.117} & \textbf{0.157}\tiny{$\pm$0.103} & \textbf{37.23} \\
& DAPS (code) & \underline{27.52} \tiny{$\pm$3.57} & \underline{0.840} \tiny{$\pm$0.084} & \underline{0.157}\tiny{$\pm$0.064} & \underline{39.39} & \textbf{26.50}\tiny{$\pm$4.70} & \underline{0.812}\tiny{$\pm$0.141} & \underline{0.170}\tiny{$\pm$0.113} & \underline{46.35} \\
& DPS & 22.73\tiny{$\pm$6.07} & 0.591\tiny{$\pm$0.141} & 0.264\tiny{$\pm$0.156} & 112.82 & 19.23\tiny{$\pm$2.52} & 0.582\tiny{$\pm$0.082} & 0.503\tiny{$\pm$0.106} & 146.23 \\
& RED-diff & 22.16\tiny{$\pm$3.41} & 0.512\tiny{$\pm$0.083} & 0.258\tiny{$\pm$0.089} & 108.32 & 22.03\tiny{$\pm$5.90} & 0.601\tiny{$\pm$0.094} & 0.274 \tiny{$\pm$0.198} & 113.48 \\

 \bottomrule
\end{tabular}%
}
\end{table*}
}

\newcommand{\tuningParams}{
\begin{table*}[]
\centering
\caption{Tuning parameters for all the linear and nonlinear tasks. Same set of parameters was used for both FFHQ and Imagenet.}
\label{tab:params}
\resizebox{0.5\textwidth}{!}{%
\begin{tabular}{@{}lccccccccc@{}}
\toprule
Task & $\eta$  & $\lambda$ & N & $\sigma_{\text{(RES)}}$ \\
 \midrule
Super Resolution &  $1.18 \times 10^{-3}$ & 11.60 & 20 & 10 \\
\midrule
Inpaint (Box) &  $4.8 \times 10^{-2}$ & 4.14 & 20  & 50 \\
 \midrule
Inpaint (Random) &  $1.1 \times 10^{-2}$ & 2.7 & 10  & 2 \\
 \midrule
Gaussian deblurring & $2.0 \times 10^{-2}$ & 0.75 & 10  & 50 \\
\midrule
Motion deblur & $2.0 \times 10^{-2}$ & 0.9 & 10  & 2 \\
\midrule
Phase Retrieval & $5.0 \times 10^{-3}$ & 0.6 & 20  & 10 \\
\midrule
Nonlinear deblur & $6.3 \times 10^{-3}$ & 2.42 & 20 & 25 \\
\midrule
High dynamic range & $3.0 \times 10^{-2}$ & 5.0 & 20  & 25\\
 \bottomrule
\end{tabular}%
}
\end{table*}
}

\newcommand{\DAPScodevspaper}{
\begin{table*}[t]
\centering
\caption{Comparison of quantitative values obtained by running DAPS official code with the values reported in DAPS paper.}
\label{tab:supp_results_daps}
\resizebox{0.9\textwidth}{!}{%
\begin{tabular}{@{}llcccccccc@{}}
\toprule
\multicolumn{1}{c}{Task} &  \multicolumn{1}{c}{Method} & \multicolumn{4}{c}{FFHQ} & \multicolumn{4}{c}{ImageNet} \\
\cmidrule(lr){3-6} \cmidrule(lr){7-10} 
 & & PSNR ($\uparrow$) & SSIM ($\uparrow$) & LPIPS ($\downarrow$) & FID ($\downarrow$) & PSNR ($\uparrow$) & SSIM ($\uparrow$) & LPIPS ($\downarrow$) & FID ($\downarrow$) \\ 
 
\midrule
\multirow{2}{*}{Super Resolution 4$\times$} 
& RePS (ours) & 29.95 & 0.845 & 0.155 & 48.37  & 26.12 & 0.708 & 0.259 & 114.19 \\

& DAPS (code) & 29.55 & 0.793 & 0.186 & 52.95  & 25.70 & 0.654 & 0.285 & 106.50  \\

& DAPS (paper) & 29.07 & 0.818 & 0.177 & 51.44  &  25.89 & 0.694 &  0.276 & 83.57 \\

\midrule
\multirow{2}{*}{Inpaint (Box)} 

& RePS (ours) & 24.32 & 0.844 & 0.122 & 41.20 & 20.71 & 0.787 & 0.186 & 109.21 \\
& DAPS (code) & 24.88 & 0.755 & 0.174 & 49.83  & 21.40 & 0.721 & 0.228 & 113.96 \\
& DAPS (paper) & 24.07 & 0.814 & 0.133 & 43.10  & 21.43 & 0.725 & 0.214 & 109.85 \\

\midrule
\multirow{1}{*}{Inpaint (Random)}  

& RePS (ours) & 32.49 & 0.899 & 0.090 & 24.18 & 28.97 & 0.829 & 0.115 & 27.52 \\
&  DAPS (code) & 30.94 & 0.807 & 0.155 & 34.84  & 27.72 & 0.742 & 0.177 & 37.18 \\
& DAPS (paper) &  31.12 &  0.844 &  0.098 &  32.17 & 28.44 & 0.775 & 0.135 & 54.25 \\

\midrule
\multirow{1}{*}{Gaussian deblurring} 

& RePS (ours) & 29.92 & 0.841 & 0.146 & 49.68 & 26.21 & 0.703 & 0.255 & 116.84 \\
& DAPS (code) & 29.75 & 0.794 & 0.177 & 53.01 & 26.00 & 0.668 & 0.260 & 102.46 \\
& DAPS (paper) & 29.19 & 0.817 & 0.165 & 53.33 & 26.15 & 0.684 & 0.253 & 75.68 \\

\midrule
\multirow{1}{*}{Motion deblurring}
& RePS (ours) & 32.27 & 0.875 & 0.115 & 34.10 & 28.95 & 0.801 & 0.169 & 53.15 \\

& DAPS (code) & 31.80 & 0.843 & 0.135 & 40.82  & 28.87 & 0.781 & 0.171 & 50.35 \\

& DAPS (paper) & 29.66 & 0.847 & 0.157 & 39.49 & 27.86 & 0.766 & 0.196 & 61.83 \\

\midrule
\multirow{1}{*}{Phase Retrieval} 

& RePS (ours)  & 30.41\tiny{$\pm$4.45} & 0.824\tiny{$\pm$0.119} & 0.152\tiny{$\pm$0.107} &  41.81 & 20.12\tiny{$\pm$7.18} & 0.449\tiny{$\pm$0.260} & 0.419\tiny{$\pm$0.184} & 220.93 \\

& DAPS (code) & 30.72\tiny{$\pm$3.15} & 0.809\tiny{$\pm$0.081} & 0.157\tiny{$\pm$0.073} & 37.88 & 22.32\tiny{$\pm$6.51} & 0.514\tiny{$\pm$0.219} & 0.343\tiny{$\pm$0.155} & 134.55 \\

& DAPS (paper) & 30.63\tiny{$\pm$3.13} &  0.851\tiny{$\pm$0.072} &  0.139\tiny{$\pm$0.060} & 42.71 & 25.78\tiny{$\pm$6.02} & 0.743\tiny{$\pm$0.084} & 0.254\tiny{$\pm$0.125} & 82.67 \\

\midrule
\multirow{1}{*}{Nonlinear deblur} 

& RePS (ours)  & 29.02\tiny{$\pm$1.76} & 0.797\tiny{$\pm$0.031} & 0.165\tiny{$\pm$0.030} & 51.97 & 27.58\tiny{$\pm$3.28} & 0.745\tiny{$\pm$0.082} & 0.191\tiny{$\pm$0.056} & 72.66 \\

& DAPS (code) & 28.79\tiny{$\pm$1.54} & 0.781\tiny{$\pm$0.033} & 0.177\tiny{$\pm$0.030} & 51.08 & 27.47\tiny{$\pm$3.21} & 0.737\tiny{$\pm$0.083} & 0.198\tiny{$\pm$0.052} & 67.62 \\

& DAPS (paper) & 28.29\tiny{$\pm$1.77} & 0.783\tiny{$\pm$0.036} & 0.155\tiny{$\pm$0.032} & 49.38 & 27.73\tiny{$\pm$3.23} & 0.724\tiny{$\pm$0.048} & 0.169\tiny{$\pm$0.056} & 59.87 \\

\midrule
\multirow{1}{*}{High dynamic range} 

& RePS (ours)  & 27.96\tiny{$\pm$3.54} & 0.872\tiny{$\pm$0.080} & 0.145\tiny{$\pm$0.071} & 36.01 & 26.37\tiny{$\pm$4.05} & 0.843\tiny{$\pm$0.117} & 0.157\tiny{$\pm$0.103} & 37.23 \\
& DAPS (code) & 27.52 \tiny{$\pm$3.57} & 0.840 \tiny{$\pm$0.084} & 0.157\tiny{$\pm$0.064} & 39.39 & 26.50\tiny{$\pm$4.70} & 0.812\tiny{$\pm$0.141} & 0.170\tiny{$\pm$0.113} & 46.35 \\
& DAPS (paper) & 27.12\tiny{$\pm$3.53} & 0.752\tiny{$\pm$0.041} & 0.162\tiny{$\pm$0.072} & 42.97 & 26.30\tiny{$\pm$4.10} & 0.717\tiny{$\pm$0.067} & 0.175\tiny{$\pm$0.107} & 64.19 \\

 \bottomrule
\end{tabular}%
}
\end{table*}
}

\definecolor{cvprblue}{rgb}{0.21,0.49,0.74}
\usepackage[pagebackref,breaklinks,colorlinks,allcolors=cvprblue]{hyperref}

\def\paperID{21241} 
\def\confName{CVPR}
\def\confYear{2026}

\rvmacroize[!]{wiener}
\rvmacroize[!]{revwiener}

\def\obsvsym{y}
\rvmacroize[][][\argcolor]{argobsv}

\def\ltntsym{x}
\rvmacroize[0]{dataltnt}
\rvmacroize[!]{recogltnt}
\rvmacroize[!]{generltnt}
\rvmacroize[!][][\argcolor]{argltnt}

\def\auxsym{z}
\rvmacroize[!]{generaux}

\rvmacroize[*]{noise}

\newcommand{\msmt}[1]{\mathbf{h}\left(#1\right)}

\def\datadistrvar{p_\text{data}}
\renewcommand{\recogmark}[1]{#1}

\pgfkeys{
    /distributions/recog/.append style = {
        index=t,  patent=\argobsvs,
    },
}
\pgfkeys{
    /distributions/gener/.append style = {
        patent=\argobsvs, 
    },
}
\pgfkeys{
    /distributions/data/.append style = {
        patent=\argobsvs, index=0
    },
}

\def\score#1 {{%
    \assignkeys{distributions, gener, spread=\mmathcolor{\argcolor}{\sigma}, adjust, #1}%
    \mathbf{s}_{\parameters}\left(\latent,\spread\right)
}}

\title{Solving Diffusion Inverse Problems with Restart Posterior Sampling}

\author{
Bilal Ahmed \qquad Joseph G.~Makin \\
Purdue University \\
{\tt\small \{ahmedb, jgmakin\}@purdue.edu}
}

\begin{document}
\maketitle
\begin{abstract}
Inverse problems are fundamental to science and engineering, where the goal is to infer an underlying signal or state from incomplete or noisy measurements. Recent approaches employ diffusion models as powerful implicit priors for such problems, owing to their ability to capture complex data distributions. However, existing diffusion-based methods for inverse problems often rely on strong approximations of the posterior distribution, require computationally expensive gradient backpropagation through the score network, or are restricted to linear measurement models. 

In this work, we propose Restart for Posterior Sampling (RePS), a general and efficient framework for solving both linear and non-linear inverse problems using pre-trained diffusion models. RePS builds on the idea of restart-based sampling, previously shown to improve sample quality in unconditional diffusion, and extends it to posterior inference. Our method employs a conditioned ODE applicable to any differentiable measurement model and introduces a simplified restart strategy that contracts accumulated approximation errors during sampling. Unlike some of the prior approaches, RePS avoids backpropagation through the score network, substantially reducing computational cost.

We demonstrate that RePS achieves faster convergence and superior reconstruction quality compared to existing diffusion-based baselines across a range of inverse problems, including both linear and non-linear settings.
\end{abstract}


\section{Introduction}
\label{sec:introduction}
Inverse problems are ubiquitous in science and engineering, arising in medical imaging, computational photography, and numerous other domains where the goal is to infer an underlying state from indirect or incomplete measurements. In most scientific applications, we have access only to partial observations that are related to the true system through a forward measurement process. The challenge lies in recovering the original signal or state from these incomplete or degraded measurements.

Examples of such settings are abundant:\ neural spiking activity reflects underlying perceptual or motor states; the number of photons captured by an image sensor corresponds to a physical scene we seek to reconstruct; and MRI measurements encode the tissue and structural properties of the human body. Despite their wide applicability, inverse problems are often difficult to solve because the observations are typically incomplete, degraded, or corrupted by measurement noise.
Consequently, the recovery problem, i.e., estimating the underlying signal from the observations, is generally ill-posed and requires incorporating prior knowledge about the solution space.
In the probabilistic framework, the measurement is modeled by a likelihood function, $\generemission{patent=\dataobsvs} $, and the constraint by a prior distribution, $\generprior{} $.
Full inversion amounts to computing the posterior distribution, $\generposterior{patent=\dataobsvs} $.

For practical problems where the solution lies in a complex, high-dimensional space, specifying an explicit prior distribution is nontrivial.
Historically, simple analytical priors, such as smoothness constraints (e.g., total-variation regularization or $\ell_2$ penalties), were employed to regularize the solution.
With the advent of data-driven methods, however, learned models have become a powerful alternative for capturing more realistic image or signal priors.
For instance, in computational imaging, \cite{venkatakrishnan_plug-and-play_2013} introduced the idea of replacing the proximal operator in iterative reconstruction with a learned denoiser.
Following this, many groups \cite{romano_little_2017, song_solving_2022, chung_solving_2023, chung_score-based_2022, kadkhodaie_solving_2021, kadkhodaie_stochastic_nodate} have attempted to regularize inverse problems with \emph{implicit} priors based on latent-variable generative models, like variational autoencoders (VAEs), generative adversarial networks (GANs), and diffusion models.

Diffusion models have achieved remarkable success as generative models for complex, high-dimensional, data distributions~\cite{ho_denoising_2020, song_score-based_2021,kim_guided-tts_nodate,iashchenko_undiff_2023, popov_grad-tts_2021,nie_scaling_2025, nie_large_2025, shi_simplified_2025} such as high-resolution images, audio, and protein structures, and are now being explored for discrete domains as well.
Their state-of-the-art sample quality, coupled with stable training dynamics, makes them highly attractive as priors for solving inverse problems.
Nevertheless, there is a major obstacle to integrating diffusion priors with measurement models:
The variables ($\generltnts{0}$) that are explicitly constrained by the measurement model (the likelihood function) are only implicitly constrained by the diffusion model (the prior)---either in the form of samples or by way of numerical integration, both computationally expensive.

One way to circumvent this obstacle is simply to introduce a third model.
For example, we might model the posterior directly by training a conditional diffusion model, e.g.\ as in classifier-free guidance.
Alternatively, if we first train an auxiliary image classifier $\generemission{index=t} $ on pairs of \emph{noisy} images $\recogltnts{t}$ and their labels $\dataobsvs$, the likelihood can be incorporated into \emph{every} step of the reverse-diffusion process \cite{dhariwal_diffusion_2021}.
However, although effective, training a separate model for every task is inefficient, and in some cases highly impractical.
Here, then, we are interested only in the ``plug-and-play'' framework that uses a pre-trained, unconditional diffusion model and a fixed measurement function, and can adapt them to arbitrary inverse problems (on the same dataset).

Crudely, there are two basic approaches.
The first is to separate the optimization scheme from the sampling process, and apply the constraint only in $\generltnts{0}$ space.
For example, after a forward pass of (unconditioned) diffusion, the measurement constraint can be imposed in $\generltnts{0}$ space with Langevin dynamics \cite{zhang_improving_2025} or gradient descent \cite{li_decoupled_2025}.
This avoids backpropagation through the score network, but risks trapping the solution in a bad local mode.
To address this, these approaches then add noise to the sample, pushing it out of the local mode.
The entire process is then repeated, with steadily decreasing noise.
This approach has been shown to recover high-quality samples.
Nevertheless, it is intuitively inefficient, since the reverse diffusions (e.g., via numerical integration of the ODE or SDE) are blind to the constraint.

The second basic approach is to attempt to optimize the likelihood and the prior simultaneously.
Since their spaces of operation are connected through the reverse-diffusion neural network, most techniques in this category backpropagate gradients through it.
For example, \citet{graikos_diffusion_2023} propose a variational-inference scheme, under which the optimization is carried out in $\generltnts{0}$ space, but which is applied to the diffusion model by making a round trip backwards through the reverse diffusion process and thence through the forward diffusion.
In DPS \cite{chung_diffusion_2024}, the measurement operator is ``lifted'' to noisy samples through the diffusion network, and again optimization requires backpropagating through it.

In this study, we propose an algorithm of the second kind (optimizing likelihood and prior simultaneously), but without backpropagating through the diffusion network.
In particular, we exploit the fact that the popular DDIM sampler for diffusion models \cite{song_denoising_2022} can be used to express a single step of an ODE simulation---from $\generltnts{t}$ to $\generltnts{t-1}$---in terms of the posterior mode, $\xpct{}{\Generltnts{0}\middle|\generltnts{t},\dataobsvs}$.
The optimal $\generltnts{0}$ is the one that lies close to the unconditional mode $\xpct{}{\Generltnts{0}\middle|\generltnts{t}}$ but also respects the measurement.
Finding this optimal value (by e.g.\ gradient descent) does not require backpropagating through the diffusion network because it answers the conditional question, ``What image $\generltnts{0}$ looks most like a denoised version of the current sample $\generltnts{t}$, subject to the constraint?''\ rather than the unconditional question, ``What sample is most congruent with this denoiser, subject to the constraint?''

Still, the method requires approximations.
Here, then, we import some observations from a recent investigation of the quality of \emph{unconditioned} diffusion sampling under various numerical-integration schemes.
In particular, \citet{xu_restart_nodate} showed that, when operating with a limited budget of neural function evaluations (NFEs), ODEs typically outperform SDEs, because they suffer from smaller discretization errors. 
On the other hand, given a large budget of NFEs, SDEs outperform ODEs, because they accumulate approximation errors---e.g., from an imperfect score function---more slowly.
Crucially, however, \cite{xu_restart_nodate} also showed how to achieve the best of both worlds, by numerically integrating an ODE (for small discretization error), but periodically ``restarting'' it, by adding a large amount of noise, in order to obliterate approximation error.

Here we apply this technique to DDIM-based sampler for inverse problems.
The resulting algorithm, which we call ``restart posterior sampling'' (RePS), extends the restart sampling concept to posterior inference. The key contributions of our method are:
\begin{itemize}
    \item{Inspired by \cite{li_efficient_2025,zhu_denoising_2023}, RePS employs an approach to condition the deterministic sampler without back-propagating through the score network.
    Unlike other recent, backprop-free methods shown to work for linear inverse problems \cite{li_efficient_2025,zhu_denoising_2023}, this method can handle \emph{any differentiable measurement model}, encompassing both linear and nonlinear cases.}
    \item{RePS incorporates a simplified restart strategy that triggers restarts only when the sample reaches the noise-free regime. Specifically, sampling is periodically restarted from $\sigma_{0}$ to a higher noise level $\sigma_r$, with $\sigma_r$ gradually annealed from a tuned maximum value down to $\sigma_{\min}$.
    }
    \item{RePS achieves faster convergence and better quality samples compared to existing baseline methods.}
\end{itemize}



\section{Background}
\label{sec:background}

\subsection{Diffusion models}
Generative diffusion models are defined in terms of a stochastic process that transforms a complicated data distribution
$\dataprior{} $
into structureless noise
$\recogprior{index=T} $.
This process can be described by a stochastic differential equation (SDE), perhaps the most common of which is the ``variance-exploding'' version:
\begin{equation}\label{eqn:forward-sde}
    \dfrntl{\recogltnts{t}}
        =
    \sqrt{2\dot\sigma(t)\sigma(t)}\, \dfrntl{\wieners{t}},
\end{equation}
where $\sigma: \mathbb{R}_{\geq 0} \to \mathbb{R}_{\geq 0}$ is a predefined noise schedule and $\wieners{} \in \mathbb{R}^d$ is a standard Wiener process.
The noise schedule is defined such that $\sigma(0)=0$ and $\sigma(T)$ is large enough to destroy any structure in the data.
It follows from \eqn{forward-sde} that the noise-perturbed distribution at step $t$, $\recogposterior{patent=\argltnts{0}} $, takes the form $\nrml{\argltnts{0}}{\sigma(t)^2\mat{I}}$.
The goal of the generative model is to invert the forward diffusion process; i.e., to start with a sample
$\generltnts{T} \sim \nrml{\vect{0}}{\sigma(T)^2\mat{I}}$
and denoise it all the way back to
$\generltnts{0} \sim \dataprior{} $.
Indeed, it is known that the forward diffusion process in \eqn{forward-sde} is invertible, with the marginal distribution over observations $\generltnts{0}$ given \cite{song_score-based_2021,karras_elucidating_2022} either by the SDE
\begin{equation}\label{eqn:reverse-sde}
    \dfrntl{\recogltnts{t}}
        =
    -2\dot{\sigma}(t)\sigma(t)
    \nabla_{\argltnts{t}}\log\recogprior{index=t} \dfrntl{t}
        +
    \sqrt{2\dot\sigma(t)\sigma(t)}\dfrntl{\revwieners{t}} ,
\end{equation}
with $\revwieners{t}$ is a reverse-time Wiener process;
or by the ODE
\begin{equation}\label{eqn:reverse-ode}
    \dfrntl{\recogltnts{t}}
        =
    -\dot\sigma(t)\sigma(t)
    \nabla_{\argltnts{t}}\log\recogprior{index=t} \dfrntl{t}.
\end{equation}
Evidently, to generate samples, we need estimates of the ``score,'' i.e.\
$\nabla_{\argltnts{t}}\log\recogprior{index=t} $,
at all $\argltnts{t}$.
This can be achieved by training a neural network $\score{} $ to match $\nabla_{\argltnts{t}}\log\recogprior{index=t} $ approximately at all noise levels $\sigma(t)$ \citep{song_generative_nodate,vincent_connection_2011}.



\schematic

\subsection{Inverse problems}
In the setting of inverse problems, the signal of interest $\dataltnts$ is not directly observed; rather, we have only measurements (or observe only degraded signals), $\dataobsvs$.
On the other hand, the relationship between the signal and its measurement is taken to be known or well described by a model.
The aim, then, is to ``invert'' this model to recover $\dataltnts$ from $\dataobsvs$.
For additive Gaussian noise, measurement model is defined as:
\begin{equation}\label{eqn:measurement-model}
    \dataobsvs = \msmt{\dataltnts} + \noises, 
\end{equation}
where $\dataltnts \in \mathbb{R}^{n}$, $\dataobsvs \in \mathbb{R}^{m}$ and $\msmt{\cdot}$: $\mathbb{R}^n \to \mathbb{R}^m$ is the measurement (or degradation) operator, which is in general nonlinear.
The term $\noises \sim \nrml{\vect{0}}{\sigma_{n}^{2}\mat{I}}$ denotes additive measurement noise. 
Such problems are often ill-posed: solutions may fail to exist, may not be unique, or may be unstable with respect to measurement noise.

To address this, prior knowledge about the solution is imposed, typically through a Bayesian formulation.
By Bayes’ rule, the posterior distribution can be written as
$\generposterior{index=0} \propto \generprior{index=0} \generemission{index=0} $.
Therefore, given the prior distribution, $\generprior{} $, and measurement model, $\generemission{} $, one can (at least in theory) compute or sample from the posterior distribution, $\generposterior{} $, to generate solutions.
In particular, we can use \eqn{reverse-sde} or \eqn{reverse-ode} to generate samples from the posterior, with the score of the posterior distribution,
\begin{equation}\label{eqn:post-score}
    \nabla_{\argltnts{t}} \log \generposterior{index=t} 
        =
    \nabla_{\argltnts{t}} \log \generprior{index=t} 
        +
    \nabla_{\argltnts{t}} \log \generemission{index=t} ,
\end{equation}
in place of the unconditional score---provided we can compute it.
The unconditional score $\nabla_{\argltnts{t}} \log \generprior{index=t} $ is (by hypothesis) available in the form of a diffusion model.
But the quantity $\nabla_{\argltnts{t}} \log \generemission{index=t} $ relates the measurement to \emph{noisy} observations $\argltnts{t}$, and is not generically computable in closed form from the likelihood score,
$\nabla_{\argltnts{0}} \log \generemission{index=0} $.


\subsection{Related work}

Several methods have been proposed to use score-based diffusion models for conditional generation in a plug-and-play fashion.
Most of these approaches aim to guide the diffusion sampling process by incorporating measurement constraints.
If we are willing to restrict ourselves to linear problems, then the singular-value decomposition \cite{kawar_denoising_2022} or projections onto the measurement space \cite{chung_score-based_2022} can be used to enforce the constraints.
Alternatively, a popular line of research extends these approaches to non-linear problems by approximating the conditional score $\generemission{index=t} $ in terms of the likelihood score $\generemission{index=0} $.
Methods such as DPS \cite{chung_diffusion_2024} use point estimates, while others employ simple unimodal distributions with estimated variances\cite{song_loss-guided_nodate}.
A major drawback of this class of methods is their computational overhead, as they require evaluating the Jacobian of the score function.

\citet{wang_dmplug_nodate} reformulate the problem from optimizing directly over $\generltnts{0}$ to optimizing the noise initialization $\generltnts{T}$, using an ODE as a deterministic mapping between these spaces.
However, this method involves backpropagating through multiple score network evaluations, which can be computationally expensive.

Variational inference, a standard approach when the posterior is not available in closed form, approximates the posterior distribution with a simpler family of distributions.
VI has recently been applied to inverse problems with diffusion priors, achieving some high-quality results \cite{graikos_diffusion_2023,mardani_variational_2023}.
However, the objective, which is computed with a sample average under the forward diffusion process, turns out to be very sensitive to which noise levels are sampled, and at which gradient step.
Very careful tuning is required to achieve good results.

A very different approach idea uses optimal-control strategies to guide the diffusion sampler.
\citet{li_solving_nodate} demonstrated promising results, but their experiments were limited to linear problems and required a large number of function evaluations (up to 2500 NFEs for linear tasks).

For unconditional diffusion sampling, \citet{xu_restart_nodate} showed that, under NFE regimes, SDE-based samplers outperform ODE-based samplers, as they more effectively contract accumulated approximation errors.
Building on this, \citet{li_efficient_2025} demonstrated that a DDIM-like sampler can be formulated as either an ODE or SDE sampler, particularly for linear inverse problems, and showed that conditioned SDE sampling produces higher-quality samples than conditioned ODE sampling by mitigating accumulated approximation errors.
We discuss the approach of \citet{li_efficient_2025} in greater detail below.

\citet{xu_restart_nodate} also showed for unconditional sampling that ODE based sampling with intermittent restarts, where large amounts of noise are added only after many steps rather than small amounts at every step as in SDE sampling, outperforms both pure ODE and pure SDE formulations.
This approach combines the low discretization error of ODEs with the noise injection benefits of SDEs, and is the inspiration for our own restart procedure.
They did not, however, provide a simple restart strategy; instead, they employed restarts at multiple noise levels along the trajectory, often repeating several restarts at each level.





Recently, a successful line of work has decoupled diffusion sampling from conditioning, using an unconditional diffusion sampler followed by conditioning in the $\generltnts{0}$ space, and then propagating back to the next (lower) noise level using the forward diffusion process \cite{li_decoupled_2025,zhang_improving_2025}.
DCDP \cite{li_decoupled_2025} enforces the constrain in $\generltnts{0}$ space via gradient descent; whereas DAPS samples from the posterior with Langevin dynamics.
DAPS, in particular, is the current state of the art \cite{zhang_improving_2025}.

\paragraph{Re-interpreting recent models in light of RePS.}  
DAPS \cite{zhang_improving_2025} and DCDP \cite{li_decoupled_2025} can be described as following three‑stages:\ an unconditioned ODE, followed by a conditioning stage, followed by forward diffusion to the next (lower) noise level.
The motivation behind these approaches was to decouple measurement conditioning from denoising.
Nevertheless, we can re-interpret them as variants on RePS. 
In particular, the second stage (addition of noise to escape local modes) can be interpreted as a restart procedure, in the manner of \citet{xu_restart_nodate}.
From this point of view, the crucial distinction between DAPS and RePS is that our algorith, replaces the two‑stage conditioning sequence (unconditioned ODE followed by data consistency) with a conditioned ODE.
By directly guiding the ODE trajectory toward the correct mode, rather than first producing an unconstrained sample that may drift toward an undesirable mode, RePS converges faster.

In contrast, DiffPIR~\cite{zhu_denoising_2023} and the algorithm of \citet{li_efficient_2025} were explicitly motivated by the notion of restart sampling.
In particular, both implement a DDIM-like sampler \cite{song_denoising_2022} that becomes deterministic when $\zeta = 0$ and stochastic for non-zero $\zeta$.
For $\alpha_t = 1$ and $\zeta = 1$, their update step can be written as
\begin{equation}
    \generltnts{t-\Delta t}
        =
    \xpct{}{\Generltnts{0} \middle| \generltnts{t},\dataobsvs} + \sigma(t-\Delta t)\rv{z},
\label{eqn:ddim-sde}
\end{equation}
where $\rv{z} \sim \mathcal{N}(\rv{0}, I)$.
Although \citet{li_efficient_2025} interpret this as a single step of an SDE sampler, it could alternatively be viewed as a single step of a conditioned ODE, followed by a restart.
Viewed in this way, their method can be seen as a special case of RePS. In contrast, RePS allows multiple conditioned ODE steps before each restart, enabling stronger performance. While these methods were demonstrated only on linear problems, RePS performs well on both linear and nonlinear problems. We discuss their relationship to RePS in the ablation results and compare performance in \cref{sec:experiments}.

\section{Methods}
\label{sec:methods}


\subsection{Measurement-conditioned ODEs}



Following \citet{karras_elucidating_2022}, we use the ODE given by \cref{eqn:reverse-ode} with noise schedule defined as $\sigma(t) = t$.
We can simulate this ODE using the posterior score to sample from the posterior distribution \cite{song_score-based_2021,chung_diffusion_2024}. 
Rewriting \cref{eqn:reverse-ode} in terms of
$\nabla_{\argltnts{t}} \log \generposterior{index=t} $
to sample from
$\generposterior{index=0} $, we obtain
\begin{equation}\label{eqn:reverse-ode-posterior}
    \dfrntl{\generltnts{t}}
        =
    -\dot\sigma(t)\sigma(t)
    \nabla_{\argltnts{t}} 
    \log\generposterior{latent=\generltnts{t},patent=\dataobsvs} 
    \dfrntl{t}.
\end{equation}
We can solve this ODE by stepping backwards in time using the Euler method:
\begin{equation}
    \generltnts{t-\Delta t}
        =
    \generltnts{t} - \Delta t\colttlderiv{\generltnts{t}}{t},
\end{equation}
where $\Delta t > 0$.
Under our noise schedule, we have
$\Delta t = \sigma(t) - \sigma(t-\Delta t)$, and the Euler update becomes
\begin{equation}
    \generltnts{t-\Delta t}
        =
    \generltnts{t}
        +
    \bigl(\sigma(t) - \sigma(t-\Delta t)\bigr)\sigma(t)
    \nabla_{\argltnts{t}} 
    \log\generposterior{latent=\generltnts{t},patent=\dataobsvs} .
\end{equation}
Following a Tweedie‑style argument (see Appendix), we obtain
\begin{equation}
    \nabla_{\argltnts{t}} \log\generposterior{latent=\generltnts{t},patent=\dataobsvs} 
        =
    \left(
        \xpct{}{\Generltnts{0} \middle|\generltnts{t},\dataobsvs} - \generltnts{t}
    \right)/\sigma^2(t),
\end{equation}
allowing us to express a measurement-conditioned numerical-integration step in terms of $\xpct{}{\Generltnts{0} \middle|\generltnts{t},\dataobsvs}$:
\begin{equation}
    \generltnts{\,t-\Delta t}
        =
    \generltnts{\,t}
        +
    \frac{\bigl(\sigma(t) - \sigma(t-\Delta t)\bigr)}{\sigma(t)}
    \left(
        \xpct{}{\Generltnts{0} \middle|\generltnts{t},\dataobsvs} - \generltnts{t}
    \right).
\label{eqn:conditioned-ode}
\end{equation}
The mean of the posterior $\generposterior{index=0,patent={\generltnts{t},\dataobsvs}} $ is generally difficult to compute, so we approximate it by the mode (MAP) of the distribution.  
Using the relation
$
\generposterior{index=0,patent={\argltnts{t},\argobsvs}} \propto
\generemission{index=0}
\generemission{index=t,patent=\argltnts{0}} 
\generprior{index=t}
$
and approximating
$\generemission{index=t,patent=\argltnts{0}} $ as
$\nrml{\xpct{}{\Generltnts{0} \middle|\generltnts{t}}}{\gamma_t^2\mat{I}}$,
the MAP estimate becomes
\begin{equation}\label{eqn:map-optimization}
    \generltnts{\rm MAP}
        =
    \argminop{\generltnts{}}{
        \frac{1}{2}
        \vectornorm{\dataobsvs - \msmt{\generltnts{}}}_2^2
            +
        \frac{\sigma_{n}^{2}}{2\gamma_t^{2}}
        \vectornorm{\generltnts{} - \xpct{}{\Generltnts{0} \middle|\generltnts{t}}}_2^2
    }.
\end{equation}

In practice, we make $\sigma_{n}^{2}/\gamma_t^2 \defeqright \lambda$ a tunable parameter, fixed for all time.
\citet{li_efficient_2025} proceed to write the solution to the optimization in \eqn{map-optimization} in closed form for linear problems---but that restricts their ODE sampler to linear or very specific non-linear problems. To keep our method general, we solve the optimization using gradient descent at each step of the numerical integration.
The resulting ODE sampler can then be used for any linear or non-linear problems with differentiable measurement operators.
Since this does not involve gradient backpropagation through the score network, the computational costs remain low.

We solve the conditioned ODE defined in \cref{eqn:conditioned-ode,eqn:map-optimization} from $\sigma(t)$ down to $\sigma_0$ using a polynomial interpolation schedule:
\begin{equation}\label{eqn:sigma-schedule}
    \sigma_s = 
    \left(
    \sigma_{\text{start}}^{\frac{1}{\rho}} 
    + s \left( 
    \sigma_{\text{end}}^{\frac{1}{\rho}} 
    - \sigma_{\text{start}}^{\frac{1}{\rho}} 
    \right)
    \right)^{\rho},
\end{equation}
where $s \in [0,1]$ is a normalized step parameter interpolating between $\sigma_{\text{start}}$ (at $s = 0$) and $\sigma_{\text{end}}$ (at $s = 1$). In all experiments in this work, we use $\rho_{\text{ode}} = 7$ for the conditioned ODE, following common practice in diffusion-based sampling \cite{karras_elucidating_2022}.


\resultsSummary


\subsection{Restart Strategy}
Instead of solving the conditioned ODEs given by \cref{eqn:conditioned-ode,eqn:map-optimization} from $\sigma_{\text{max}}$ to $\sigma_{0}$ in a single pass, we adopt a restart-based approach as shown in \cref{fig:schematic}. Specifically, we first solve the ODE for a small number of steps down to $\sigma_{0}$, then perform a forward diffusion step to reintroduce noise at a predefined restart level $\sigma_{\text{restart}}$. From this point, we again solve the ODE backward to $\sigma_{0}$. This process is repeated multiple times, with the restart level gradually annealed from $\sigma_{\text{restart}}$ to $\sigma_{\text{min}}$.

Since we employ the VE‑SDE formulation, each restart step is defined as:
\begin{equation}\label{eqn:restart‑step}
    \generltnts{r}
        \sim
    \nrml{\generltnts{0}}{\sigma_r^2\mat{I}},
\end{equation}
where $\sigma_r$ denotes the restart noise level.
\citet{xu_restart_nodate} described a restart scheme that injected noise at multiple arbitrary levels and repeated each level several times. In contrast, our strategy replaces that complexity with a much simpler schedule, allowing us to employ restart‑based sampling as an effective alternative to standard SDE or ODE approaches.

In all our experiments, we adopt a simple restart schedule based on the very same polynomial interpolation used for the ODE simulator, namely, \eqn{sigma-schedule}.
Specifically, the restart schedule interpolates from the maximum restart level $\sigma_{\mathrm{restart}}$ to the minimum restart level $\sigma_{\mathrm{min}}$. We find that concentrating restart iterations at smaller noise levels improves performance. Accordingly, we tune $\sigma_{\mathrm{restart}}$ separately for each task, while keeping $\sigma_{\mathrm{min}}$ fixed.



The full sampling procedure is summarized in \cref{alg:reps} in the Appendix.


\section{Experiments}
\label{sec:experiments}

We evaluate our method on two image datasets: ImageNet 256×256 \cite{deng_imagenet_2009} and FFHQ 256×256 \cite{karras_style-based_2019}. For ImageNet, we use the pretrained diffusion model publicly released by Dhariwal and Nichol \cite{dhariwal_diffusion_2021}, and for FFHQ, we use the pretrained model provided by Chung et al. \cite{chung_diffusion_2024}. Although both of these models were trained using DDPM framework that corresponds to VP-SDE, following \cite{karras_elucidating_2022} we use the unified VE-SDE based sampler for these models.
We consider a range of linear and non-linear inverse problems that are widely studied in the literature on image restoration and reconstruction.
For all experiments, we begin from a noise level of $\sigma_{\text{max}} = 100$ and use 10 steps of the conditioned ODE to move from the starting noise level (or the current restart noise level) down to $\sigma_0 = 0.01$, following the polynomial schedule defined in \cref{eqn:sigma-schedule} with $\rho_{\text{ode}} = 7$.  
The restart schedule is then applied, beginning at task‑specific $\sigma_{\text{restart}}$ and interpolating down to a minimum restart level $\sigma_{\min}=0.1$, using the same polynomial form with $\rho_{\text{restart}} = 15$.  
In addition to $\sigma_{\text{restart}}$, we tune the number of gradient updates per step \(N\), the learning rate \(\eta\), and the parameter \(\lambda\) for each task; these are summarized in \cref{sec:exp-details}. 
\paragraph{Linear problems.} 
The linear inverse problems considered in our experiments include:
(1) super-resolution,
(2) inpainting (box mask),
(3) inpainting (random mask),
(4) Gaussian deblurring, and
(5) motion deblurring.

In super-resolution, the goal is to recover a high-resolution image from a low-resolution observation obtained by bicubic downsampling with a factor of 4.
In box inpainting, each image is masked by a randomly placed box of size 128×128, while in random inpainting, 70\% of the pixels are randomly masked.
For Gaussian deblurring, we use a blur kernel of size 61×61 with a standard deviation of 3.0, and for motion deblurring, we use a kernel of the same size with a standard deviation of 0.5.
In all tasks, Gaussian noise with a standard deviation of $\sigma_n = 0.05$ is added to the measurements.
\paragraph{Non-linear problems.}
The non-linear inverse problems considered include:
(1) phase retrieval,
(2) non-linear deblurring, and
(3) high dynamic range (HDR) reconstruction.

In phase retrieval, the objective is to reconstruct an image from only the magnitude of its Fourier transform, where the phase information is missing. Since this is a highly ill-posed problem, we follow previous work \cite{chung_diffusion_2024, zhang_improving_2025} in using measurements oversampled by a factor of 2, generating 4 reconstructions per measurement, and reporting the best score among them.

Non-linear deblurring involves removing blur introduced by a neural network. For this task, we use the pretrained model publicly released by \citet{tran_explore_2021}.

In HDR, the goal is to recover the full dynamic range of pixel intensities from measurements in which pixel values are compressed to a smaller range by a factor of 2.
As in the linear problems, Gaussian noise with $\sigma_n = 0.05$ is added to the measurements.

\paragraph{Baselines and metrics.}
We compare our method against the following baselines:
\citet{li_efficient_2025},
DAPS~\cite{zhang_improving_2025},
DDRM~\cite{kawar_denoising_2022},
DDNM~\cite{wang_zero-shot_2022},
DPS~\cite{chung_diffusion_2024},
DCDP~\cite{li_decoupled_2025},
FPS-SMC~\cite{dou_diffusion_2024}, and
DiffPIR~\cite{zhu_denoising_2023} for linear problems; and DAPS~\cite{zhang_improving_2025}, DPS~\cite{chung_diffusion_2024},
RED-diff~\cite{mardani_variational_2023},
and DCDP~\cite{li_decoupled_2025} for non-linear problems.
To quantitatively evaluate performance, we report four standard metrics: peak signal-to-noise ratio (PSNR), structural similarity index measure (SSIM), learned perceptual image patch similarity (LPIPS), and Fréchet inception distance (FID).
FID measures the fidelity of the generated samples with respect to the true data distribution, whereas PSNR, SSIM, and LPIPS assess the similarity between the generated and ground-truth images, thereby reflecting the degree of measurement consistency.

Since we use the same test setup as DAPS, most baseline results are reported from their paper. However, results obtained by running the DAPS code did not exactly match the reported values, so we report the numbers from running their official code. For comparison, we include their original results from the paper in the supplementary material. \citet{li_efficient_2025} did not provide code, so we implemented their method ourselves; we report their results for 1000 NFEs, using gradient descent rather than the closed-form solution.

\subsection{Main Results}
In 
\cref{fig:qualitative},
we present qualitative visualisations for both linear and non‑linear inverse problems. These results show that RePS consistently recovers fine structural and textural details across a variety of tasks, including phase‑retrieval and non‑linear deblur.  
In
\cref{fig:multi-runs},
we show multiple reconstructions of the same input: for phase‑retrieval, RePS produces high‑fidelity results reliably; for inpainting, it generates diverse yet plausible outputs in the masked regions. 
We also include generated examples for all linear and non‑linear tasks in \cref{sec:add-results}.

Quantitative comparisons of RePS with baseline methods are provided in \cref{tab:main_results}.  
RePS outperforms the recent state-of-the-art method DAPS across all linear tasks and on two of the three nonlinear tasks.  
While individual baselines such as \citet{li_efficient_2025} for inpaint (random) and RED-diff for nonlinear deblur achieve strong performance on specific tasks, RePS nonetheless delivers the best overall results across the full range of tasks studied.

\begin{figure}[t]
  \centering
    \newcommand{\tikzsubdir}{\tikzdir/sde-ode}%
    \newcommand{\figwidth}{\linewidth}%
    \newcommand{\figheight}{0.4\linewidth}%
    \newcommand{\marksize}{10}%
    \provideboolean{CLEANTITLE}\setboolean{CLEANTITLE}{true}%

    \begin{tabular}{@{}cc@{}}
        \provideboolean{CLEANXAXIS}\setboolean{CLEANXAXIS}{true}%
\begin{tikzpicture}
\provideboolean{CLEARXLABEL}\ifthenelse{\boolean{CLEARXLABEL}}{%
	\pgfplotsset{every axis post/.append style={xlabel = {} }}%
}{}%
\providecommand{\figheight}{2.0in}%
\pgfplotsset{compat=1.15}%
\provideboolean{NOLEGEND}%
\providecommand{\thisYlabelopacity}{1.0}%
\providecommand{\thisXlabelopacity}{1.0}%
\provideboolean{CLEANXAXIS}\ifthenelse{\boolean{CLEANXAXIS}}{%
	\pgfplotsset{every axis post/.append style={xlabel = {} }}%
}{}%
\providecommand{\figwidth}{5.7in}%
\provideboolean{CLEANYAXIS}\ifthenelse{\boolean{CLEANYAXIS}}{%
	\pgfplotsset{every axis post/.append style={ylabel = {} }}%
}{}%
\providecommand{\figwidth}{360pt}%
\provideboolean{CLEANTITLE}\ifthenelse{\boolean{CLEANTITLE}}{%
	\pgfplotsset{every axis post/.append style={title = {} }}%
}{}%
\providecommand{\marksize}{2}%
\providecommand{\thisXticklabelopacity}{1.0}%
\provideboolean{CLEANXAXIS}\ifthenelse{\boolean{CLEANXAXIS}}{%
	\pgfplotsset{every axis post/.append style={xticklabels = {} }}%
}{}%
\provideboolean{CLEANYAXIS}\ifthenelse{\boolean{CLEANYAXIS}}{%
	\pgfplotsset{every axis post/.append style={yticklabels = {} }}%
}{}%
\providecommand{\figheight}{310pt}%
\provideboolean{CLEARYLABEL}\ifthenelse{\boolean{CLEARYLABEL}}{%
	\pgfplotsset{every axis post/.append style={ylabel = {} }}%
}{}%

\definecolor{darkgray176}{RGB}{176,176,176}
\definecolor{darkorange25512714}{RGB}{255,127,14}
\definecolor{forestgreen4416044}{RGB}{44,160,44}
\definecolor{gray127}{RGB}{127,127,127}
\definecolor{lightgray204}{RGB}{204,204,204}

\begin{axis}[
axis lines=left,
every axis plot/.append style={mark size=\marksize},
every axis x label/.append style={opacity=\thisXlabelopacity},
every axis y label/.append style={opacity=\thisYlabelopacity},
every x tick label/.append style={rotate=0},
height=\figheight,
legend cell align={left},
legend style={
  fill opacity=0.8,
  draw opacity=1,
  text opacity=1,
  at={(0.97,0.03)},
  anchor=south east,
  draw=lightgray204
},
mark size=\marksize,
tick align=outside,
tick pos=left,
title={FFHQ - Gaussian Deblur},
width=\figwidth,
x grid style={darkgray176},
xlabel={NFE},
xmajorgrids,
xmin=-44.75, xmax=1049.75,
xtick style={color=black},
xticklabel style={opacity=\thisXticklabelopacity, align=center},
y grid style={darkgray176},
ylabel={PSNR},
ymajorgrids,
ymin=3.68695, ymax=31.16605,
ytick style={color=black}
]
\addplot [semithick, darkorange25512714, mark=*, mark size=2, mark options={solid}]
table {%
20 25.707
50 27.926
100 28.569
200 29.191
400 29.599
600 29.729
800 29.884
1000 29.917
};
\addlegendentry{RePS}
\addplot [semithick, forestgreen4416044, mark=*, mark size=2, mark options={solid}]
table {%
5 4.936
10 5.092
20 6.192
50 10.89
100 17.687
200 23.895
400 25.743
600 26.293
800 26.495
1000 26.596
};
\addlegendentry{SDE}
\addplot [semithick, gray127, mark=*, mark size=2, mark options={solid}]
table {%
5 21.119
10 24.068
20 24.947
50 25.003
100 24.902
200 24.831
400 24.795
600 24.781
800 24.774
1000 24.681
};
\addlegendentry{ODE}
\ifthenelse{\boolean{NOLEGEND}}{\legend{}}{}
\end{axis}

\end{tikzpicture} \\
\begin{tikzpicture}
\provideboolean{CLEARXLABEL}\ifthenelse{\boolean{CLEARXLABEL}}{%
	\pgfplotsset{every axis post/.append style={xlabel = {} }}%
}{}%
\providecommand{\figheight}{2.0in}%
\pgfplotsset{compat=1.15}%
\provideboolean{NOLEGEND}%
\providecommand{\thisYlabelopacity}{1.0}%
\providecommand{\thisXlabelopacity}{1.0}%
\provideboolean{CLEANXAXIS}\ifthenelse{\boolean{CLEANXAXIS}}{%
	\pgfplotsset{every axis post/.append style={xlabel = {} }}%
}{}%
\providecommand{\figwidth}{5.7in}%
\provideboolean{CLEANYAXIS}\ifthenelse{\boolean{CLEANYAXIS}}{%
	\pgfplotsset{every axis post/.append style={ylabel = {} }}%
}{}%
\providecommand{\figwidth}{360pt}%
\provideboolean{CLEANTITLE}\ifthenelse{\boolean{CLEANTITLE}}{%
	\pgfplotsset{every axis post/.append style={title = {} }}%
}{}%
\providecommand{\marksize}{2}%
\providecommand{\thisXticklabelopacity}{1.0}%
\provideboolean{CLEANXAXIS}\ifthenelse{\boolean{CLEANXAXIS}}{%
	\pgfplotsset{every axis post/.append style={xticklabels = {} }}%
}{}%
\provideboolean{CLEANYAXIS}\ifthenelse{\boolean{CLEANYAXIS}}{%
	\pgfplotsset{every axis post/.append style={yticklabels = {} }}%
}{}%
\providecommand{\figheight}{310pt}%
\provideboolean{CLEARYLABEL}\ifthenelse{\boolean{CLEARYLABEL}}{%
	\pgfplotsset{every axis post/.append style={ylabel = {} }}%
}{}%

\definecolor{darkgray176}{RGB}{176,176,176}
\definecolor{darkorange25512714}{RGB}{255,127,14}
\definecolor{forestgreen4416044}{RGB}{44,160,44}
\definecolor{gray127}{RGB}{127,127,127}
\definecolor{lightgray204}{RGB}{204,204,204}

\begin{axis}[
axis lines=left,
every axis plot/.append style={mark size=\marksize},
every axis x label/.append style={opacity=\thisXlabelopacity},
every axis y label/.append style={opacity=\thisYlabelopacity},
every x tick label/.append style={rotate=0},
height=\figheight,
legend cell align={left},
legend style={fill opacity=0.8, draw opacity=1, text opacity=1, draw=lightgray204},
mark size=\marksize,
tick align=outside,
tick pos=left,
title={FFHQ - Gaussian Deblur},
width=\figwidth,
x grid style={darkgray176},
xlabel={NFE},
xmajorgrids,
xmin=-44.75, xmax=1049.75,
xtick style={color=black},
xticklabel style={opacity=\thisXticklabelopacity, align=center},
y grid style={darkgray176},
ylabel={LPIPS},
ymajorgrids,
ymin=0.10855, ymax=0.91045,
ytick style={color=black}
]
\addplot [semithick, darkorange25512714, mark=*, mark size=2, mark options={solid}]
table {%
20 0.267
50 0.166
100 0.155
200 0.147
400 0.145
600 0.145
800 0.146
1000 0.146
};
\addlegendentry{RePS}
\addplot [semithick, forestgreen4416044, mark=*, mark size=2, mark options={solid}]
table {%
5 0.874
10 0.864
20 0.84
50 0.754
100 0.508
200 0.286
400 0.231
600 0.214
800 0.208
1000 0.205
};
\addlegendentry{SDE}
\addplot [semithick, gray127, mark=*, mark size=2, mark options={solid}]
table {%
5 0.424
10 0.332
20 0.308
50 0.301
100 0.302
200 0.303
400 0.303
600 0.303
800 0.303
1000 0.303
};
\addlegendentry{ODE}
\ifthenelse{\boolean{NOLEGEND}}{\legend{}}{}
\end{axis}

\end{tikzpicture} \\
    \end{tabular}%

    \captioning{Comparison to ODE and SDE samplers}{We plot two metrics, PSNR (top) and LPIPS (bottom), for Gaussian deblurring on FFHQ, as a function of NFEs for three different samplers: SDE, ODE, and RePS. The plots indicate that RePS achieves faster convergence and better overall performance.}

    \label{fig:sde-comparison}
\end{figure}

\paragraph{Comparison to SDE and ODE.}
To highlight the performance gain achieved through restart sampling, we compare PSNR and LPIPS on Gaussian Deblur in \cref{fig:sde-comparison} using three approaches: (i) the same ODE without any restart, (ii) an SDE derived via the Euler–Maruyama method (details in the Appendix) with identical conditioning but continuous noise injection, and (iii) our proposed RePS method.  
The results confirm that the ODE exhibits low discretisation error and converges rapidly at low NFEs, while the SDE delivers stronger performance at higher NFEs due to the error contraction effects of stochastic noise.  
RePS combines both strengths as it converges quickly and attains the best performance across the full NFE range.

As discussed in \cref{sec:background}, \citet{zhu_denoising_2023} and \citet{li_efficient_2025} employ an SDE derived from the DDIM sampler, which can be viewed as a special case of our RePS framework---that is, restarting interpolated with ODE simulation, but using only a single ODE step.  
We present ablation results in \cref{fig:ablation} that examine the influence of the number of ODE steps taken between successive restarts (with the total NFEs held constant at 1000).  
The leftmost data point in the figure corresponds to \eqn{ddim-sde}; i.e., one conditioned ODE step between restarts.  
The results show that increasing the number of ODE steps between restarts improves performance up to a point. Beyond approximately 20 ODE steps, LPIPS begins to worsen, while PSNR peaks around 5 ODE steps. 
This result is consistent with our comparison to the standard Euler–Maruyama SDE and is also reflected in our main results in \cref{tab:main_results}, where RePS outperforms DiffPIR across all tasks and also outperforms \citet{li_efficient_2025} in all tasks except Inpaint (Random).

\paragraph{Comparison to DAPS.}  
DAPS is the current state-of-the-art method, so in addition to the results in \cref{tab:main_results} we compare RePS with DAPS at different NFEs. 
\cref{fig:daps-comparison} shows that RePS achieves better PSNR and LPIPS across all NFEs compared to DAPS on the motion-deblurring task.
For completeness, we provide results including SSIM and FID scores for all linear and nonlinear problems in \cref{sec:add-results}. 
In \cref{fig:sampling-times} we compare the per‐image sampling time of RePS and DAPS across all tasks.
The times shown correspond to the maximum NFEs used for each task (i.e., 1k for linear problems and 4k for nonlinear problems).
Overall, the sampling times are comparable: RePS is slightly faster on linear tasks and slightly slower on nonlinear tasks, but the differences are small.

\begin{figure}[t]
  \centering
    \newcommand{\tikzsubdir}{\tikzdir/reps-daps}%
    \newcommand{\figwidth}{\linewidth}%
    \newcommand{\figheight}{0.4\linewidth}%
    \newcommand{\marksize}{2.5}%
    \provideboolean{CLEANTITLE}\setboolean{CLEANTITLE}{true}%

    \begin{tabular}{@{}cc@{}}
        \provideboolean{CLEANXAXIS}\setboolean{CLEANXAXIS}{true}%
\begin{tikzpicture}
\providecommand{\figwidth}{5.7in}%
\providecommand{\thisXlabelopacity}{1.0}%
\providecommand{\marksize}{1}%
\provideboolean{CLEANXAXIS}\ifthenelse{\boolean{CLEANXAXIS}}{%
	\pgfplotsset{every axis post/.append style={xlabel = {} }}%
}{}%
\provideboolean{CLEANTITLE}\ifthenelse{\boolean{CLEANTITLE}}{%
	\pgfplotsset{every axis post/.append style={title = {} }}%
}{}%
\provideboolean{CLEANYAXIS}\ifthenelse{\boolean{CLEANYAXIS}}{%
	\pgfplotsset{every axis post/.append style={yticklabels = {} }}%
}{}%
\provideboolean{CLEARXLABEL}\ifthenelse{\boolean{CLEARXLABEL}}{%
	\pgfplotsset{every axis post/.append style={xlabel = {} }}%
}{}%
\provideboolean{CLEANYAXIS}\ifthenelse{\boolean{CLEANYAXIS}}{%
	\pgfplotsset{every axis post/.append style={ylabel = {} }}%
}{}%
\providecommand{\thisYlabelopacity}{1.0}%
\providecommand{\figwidth}{360pt}%
\provideboolean{NOLEGEND}%
\providecommand{\figheight}{310pt}%
\provideboolean{CLEARYLABEL}\ifthenelse{\boolean{CLEARYLABEL}}{%
	\pgfplotsset{every axis post/.append style={ylabel = {} }}%
}{}%
\pgfplotsset{compat=1.15}%
\providecommand{\thisXticklabelopacity}{1.0}%
\providecommand{\figheight}{2.0in}%
\provideboolean{CLEANXAXIS}\ifthenelse{\boolean{CLEANXAXIS}}{%
	\pgfplotsset{every axis post/.append style={xticklabels = {} }}%
}{}%

\definecolor{darkgray176}{RGB}{176,176,176}
\definecolor{darkorange25512714}{RGB}{255,127,14}
\definecolor{lightgray204}{RGB}{204,204,204}
\definecolor{steelblue31119180}{RGB}{31,119,180}

\begin{axis}[
axis lines=left,
every axis plot/.append style={mark size=\marksize},
every axis x label/.append style={opacity=\thisXlabelopacity},
every axis y label/.append style={opacity=\thisYlabelopacity},
every x tick label/.append style={rotate=0},
height=\figheight,
legend cell align={left},
legend style={
  fill opacity=0.8,
  draw opacity=1,
  text opacity=1,
  at={(0.97,0.03)},
  anchor=south east,
  draw=lightgray204
},
mark size=\marksize,
tick align=outside,
tick pos=left,
title={FFHQ - Motion Deblur},
width=\figwidth,
x grid style={darkgray176},
xlabel={NFEs},
xmajorgrids,
xmin=-29, xmax=1049,
xtick style={color=black},
xticklabel style={opacity=\thisXticklabelopacity, align=center},
y grid style={darkgray176},
ylabel={PSNR},
ymajorgrids,
ymin=28.41845, ymax=32.45655,
ytick style={color=black}
]
\addplot [semithick, darkorange25512714, mark=*, mark size=2, mark options={solid}]
table {%
20 28.602
50 30.354
100 31.218
200 31.811
400 32.135
600 32.214
800 32.267
1000 32.273
};
\addlegendentry{RePS}
\addplot [semithick, steelblue31119180, mark=*, mark size=2, mark options={solid}]
table {%
50 29.071
100 30.511
200 31.438
400 31.246
1000 31.796
};
\addlegendentry{DAPS}
\ifthenelse{\boolean{NOLEGEND}}{\legend{}}{}
\end{axis}

\end{tikzpicture} \\
\begin{tikzpicture}
\providecommand{\figwidth}{5.7in}%
\providecommand{\thisXlabelopacity}{1.0}%
\providecommand{\marksize}{1}%
\provideboolean{CLEANXAXIS}\ifthenelse{\boolean{CLEANXAXIS}}{%
	\pgfplotsset{every axis post/.append style={xlabel = {} }}%
}{}%
\provideboolean{CLEANTITLE}\ifthenelse{\boolean{CLEANTITLE}}{%
	\pgfplotsset{every axis post/.append style={title = {} }}%
}{}%
\provideboolean{CLEANYAXIS}\ifthenelse{\boolean{CLEANYAXIS}}{%
	\pgfplotsset{every axis post/.append style={yticklabels = {} }}%
}{}%
\provideboolean{CLEARXLABEL}\ifthenelse{\boolean{CLEARXLABEL}}{%
	\pgfplotsset{every axis post/.append style={xlabel = {} }}%
}{}%
\provideboolean{CLEANYAXIS}\ifthenelse{\boolean{CLEANYAXIS}}{%
	\pgfplotsset{every axis post/.append style={ylabel = {} }}%
}{}%
\providecommand{\thisYlabelopacity}{1.0}%
\providecommand{\figwidth}{360pt}%
\provideboolean{NOLEGEND}%
\providecommand{\figheight}{310pt}%
\provideboolean{CLEARYLABEL}\ifthenelse{\boolean{CLEARYLABEL}}{%
	\pgfplotsset{every axis post/.append style={ylabel = {} }}%
}{}%
\pgfplotsset{compat=1.15}%
\providecommand{\thisXticklabelopacity}{1.0}%
\providecommand{\figheight}{2.0in}%
\provideboolean{CLEANXAXIS}\ifthenelse{\boolean{CLEANXAXIS}}{%
	\pgfplotsset{every axis post/.append style={xticklabels = {} }}%
}{}%

\definecolor{darkgray176}{RGB}{176,176,176}
\definecolor{darkorange25512714}{RGB}{255,127,14}
\definecolor{lightgray204}{RGB}{204,204,204}
\definecolor{steelblue31119180}{RGB}{31,119,180}

\begin{axis}[
axis lines=left,
every axis plot/.append style={mark size=\marksize},
every axis x label/.append style={opacity=\thisXlabelopacity},
every axis y label/.append style={opacity=\thisYlabelopacity},
every x tick label/.append style={rotate=0},
height=\figheight,
legend cell align={left},
legend style={fill opacity=0.8, draw opacity=1, text opacity=1, draw=lightgray204},
mark size=\marksize,
tick align=outside,
tick pos=left,
title={FFHQ - Motion Deblur},
width=\figwidth,
x grid style={darkgray176},
xlabel={NFEs},
xmajorgrids,
xmin=-29, xmax=1049,
xtick style={color=black},
xticklabel style={opacity=\thisXticklabelopacity, align=center},
y grid style={darkgray176},
ylabel={LPIPS},
ymajorgrids,
ymin=0.1096, ymax=0.2284,
ytick style={color=black}
]
\addplot [semithick, darkorange25512714, mark=*, mark size=2, mark options={solid}]
table {%
20 0.171
50 0.138
100 0.128
200 0.121
400 0.117
600 0.116
800 0.115
1000 0.115
};
\addlegendentry{RePS}
\addplot [semithick, steelblue31119180, mark=*, mark size=2, mark options={solid}]
table {%
50 0.223
100 0.182
200 0.155
400 0.153
1000 0.135
};
\addlegendentry{DAPS}
\ifthenelse{\boolean{NOLEGEND}}{\legend{}}{}
\end{axis}

\end{tikzpicture} \\
    \end{tabular}%

    \captioning{Comparison to DAPS at different NFEs}{We plot PSNR and LPIPS for DAPS and RePS on FFHQ motion deblurring, as a function of NFEs. The plots show that RePS achieves better performance across different NFEs.}

    \label{fig:daps-comparison}
\end{figure}

\begin{figure}[t]
  \centering
    \newcommand{\figwidth}{\linewidth}%
    \newcommand{\figheight}{0.5\linewidth}%
    \newcommand{\marksize}{2.5}%
    \provideboolean{CLEANTITLE}\setboolean{CLEANTITLE}{true}%
\begin{tikzpicture}
\providecommand{\thisXticklabelopacity}{1.0}%
\providecommand{\figheight}{310pt}%
\provideboolean{CLEANTITLE}\ifthenelse{\boolean{CLEANTITLE}}{%
	\pgfplotsset{every axis post/.append style={title = {} }}%
}{}%
\provideboolean{CLEARXLABEL}\ifthenelse{\boolean{CLEARXLABEL}}{%
	\pgfplotsset{every axis post/.append style={xlabel = {} }}%
}{}%
\provideboolean{CLEANXAXIS}\ifthenelse{\boolean{CLEANXAXIS}}{%
	\pgfplotsset{every axis post/.append style={xlabel = {} }}%
}{}%
\provideboolean{CLEANYAXIS}\ifthenelse{\boolean{CLEANYAXIS}}{%
	\pgfplotsset{every axis post/.append style={ylabel = {} }}%
}{}%
\providecommand{\thisYlabelopacity}{1.0}%
\providecommand{\thisXlabelopacity}{1.0}%
\provideboolean{CLEANYAXIS}\ifthenelse{\boolean{CLEANYAXIS}}{%
	\pgfplotsset{every axis post/.append style={yticklabels = {} }}%
}{}%
\providecommand{\figheight}{2.0in}%
\provideboolean{NOLEGEND}%
\providecommand{\marksize}{2}%
\pgfplotsset{compat=1.15}%
\provideboolean{CLEANXAXIS}\ifthenelse{\boolean{CLEANXAXIS}}{%
	\pgfplotsset{every axis post/.append style={xticklabels = {} }}%
}{}%
\providecommand{\figwidth}{360pt}%
\providecommand{\figwidth}{5.7in}%
\provideboolean{CLEARYLABEL}\ifthenelse{\boolean{CLEARYLABEL}}{%
	\pgfplotsset{every axis post/.append style={ylabel = {} }}%
}{}%

\definecolor{darkgray176}{RGB}{176,176,176}
\definecolor{darkorange25512714}{RGB}{255,127,14}
\definecolor{lightgray204}{RGB}{204,204,204}
\definecolor{steelblue31119180}{RGB}{31,119,180}

\begin{axis}[
axis lines=left,
every axis plot/.append style={mark size=\marksize},
every axis x label/.append style={opacity=\thisXlabelopacity},
every axis y label/.append style={opacity=\thisYlabelopacity},
every x tick label/.append style={rotate=0},
height=\figheight,
legend cell align={left},
legend style={
  fill opacity=0.8,
  draw opacity=1,
  text opacity=1,
  at={(0.03,0.97)},
  anchor=north west,
  draw=lightgray204
},
mark size=\marksize,
tick align=outside,
tick pos=left,
title={Comparison of Methods Across Tasks},
width=\figwidth,
x grid style={darkgray176},
xlabel={Task},
xmin=-0.968, xmax=7.568,
xtick style={color=black},
xtick={0,1,2,3,4,5,6,7},
xticklabel style={opacity=\thisXticklabelopacity, align=center},
xticklabels={
  SR,
  IB,
  IR,
  GD,
  MD,
  PR,
  ND,
  HDR
},
y grid style={darkgray176},
ylabel={Running Time (sec)},
ymin=0, ymax=236.9115,
ytick style={color=black}
]
\draw[draw=none,fill=darkorange25512714] (axis cs:-0.58,0) rectangle (axis cs:-0.22,10.68);
\addlegendimage{ybar,ybar legend,draw=none,fill=darkorange25512714}
\addlegendentry{RePS}

\draw[draw=none,fill=darkorange25512714] (axis cs:0.42,0) rectangle (axis cs:0.78,10.11);
\draw[draw=none,fill=darkorange25512714] (axis cs:1.42,0) rectangle (axis cs:1.78,10.65);
\draw[draw=none,fill=darkorange25512714] (axis cs:2.42,0) rectangle (axis cs:2.78,27.27);
\draw[draw=none,fill=darkorange25512714] (axis cs:3.42,0) rectangle (axis cs:3.78,25.43);
\draw[draw=none,fill=darkorange25512714] (axis cs:4.42,0) rectangle (axis cs:4.78,53.48);
\draw[draw=none,fill=darkorange25512714] (axis cs:5.42,0) rectangle (axis cs:5.78,225.63);
\draw[draw=none,fill=darkorange25512714] (axis cs:6.42,0) rectangle (axis cs:6.78,39.95);
\draw[draw=none,fill=steelblue31119180] (axis cs:-0.18,0) rectangle (axis cs:0.18,11.75);
\addlegendimage{ybar,ybar legend,draw=none,fill=steelblue31119180}
\addlegendentry{DAPS}

\draw[draw=none,fill=steelblue31119180] (axis cs:0.82,0) rectangle (axis cs:1.18,10.95);
\draw[draw=none,fill=steelblue31119180] (axis cs:1.82,0) rectangle (axis cs:2.18,9.92);
\draw[draw=none,fill=steelblue31119180] (axis cs:2.82,0) rectangle (axis cs:3.18,49.29);
\draw[draw=none,fill=steelblue31119180] (axis cs:3.82,0) rectangle (axis cs:4.18,49.3);
\draw[draw=none,fill=steelblue31119180] (axis cs:4.82,0) rectangle (axis cs:5.18,44.67);
\draw[draw=none,fill=steelblue31119180] (axis cs:5.82,0) rectangle (axis cs:6.18,223.85);
\draw[draw=none,fill=steelblue31119180] (axis cs:6.82,0) rectangle (axis cs:7.18,38.07);
\ifthenelse{\boolean{NOLEGEND}}{\legend{}}{}
\end{axis}

\end{tikzpicture}

    \captioning{Sampling time comparison of RePS versus DAPS.}{The figure shows the time per sample (in seconds) across all tasks in the study: Super-resolution (SR), Inpaint-Box (IB), Inpaint-Random (IR), Gaussian-Deblur (GD), Motion-Deblur (MD), Phase-Retrieval (PR), Nonlinear-Deblur (NB) and High-Dynamic-Range (HDR).}
    \label{fig:sampling-times}
\end{figure}

\begin{figure}[t]
  \centering
    \newcommand{\figwidth}{\linewidth}%
    \newcommand{\figheight}{0.4\linewidth}%
    \newcommand{\marksize}{2.5}%
    \provideboolean{CLEANTITLE}\setboolean{CLEANTITLE}{true}%

    \begin{tabular}{@{}cc@{}}
        \provideboolean{CLEANXAXIS}\setboolean{CLEANXAXIS}{true}%
\begin{tikzpicture}
\providecommand{\thisYlabelopacity}{1.0}%
\providecommand{\figwidth}{360pt}%
\provideboolean{CLEANTITLE}\ifthenelse{\boolean{CLEANTITLE}}{%
	\pgfplotsset{every axis post/.append style={title = {} }}%
}{}%
\providecommand{\figheight}{310pt}%
\provideboolean{CLEANYAXIS}\ifthenelse{\boolean{CLEANYAXIS}}{%
	\pgfplotsset{every axis post/.append style={ylabel = {} }}%
}{}%
\providecommand{\thisXticklabelopacity}{1.0}%
\provideboolean{CLEANYAXIS}\ifthenelse{\boolean{CLEANYAXIS}}{%
	\pgfplotsset{every axis post/.append style={yticklabels = {} }}%
}{}%
\provideboolean{CLEANXAXIS}\ifthenelse{\boolean{CLEANXAXIS}}{%
	\pgfplotsset{every axis post/.append style={xticklabels = {} }}%
}{}%
\provideboolean{CLEANXAXIS}\ifthenelse{\boolean{CLEANXAXIS}}{%
	\pgfplotsset{every axis post/.append style={xlabel = {} }}%
}{}%
\providecommand{\figheight}{2.0in}%
\provideboolean{NOLEGEND}%
\pgfplotsset{compat=1.15}%
\providecommand{\figwidth}{5.7in}%
\provideboolean{CLEARYLABEL}\ifthenelse{\boolean{CLEARYLABEL}}{%
	\pgfplotsset{every axis post/.append style={ylabel = {} }}%
}{}%
\providecommand{\thisXlabelopacity}{1.0}%
\providecommand{\marksize}{1}%
\provideboolean{CLEARXLABEL}\ifthenelse{\boolean{CLEARXLABEL}}{%
	\pgfplotsset{every axis post/.append style={xlabel = {} }}%
}{}%

\definecolor{darkgray176}{RGB}{176,176,176}
\definecolor{lightgray204}{RGB}{204,204,204}
\definecolor{purple}{RGB}{128,0,128}

\begin{axis}[
axis lines=left,
every axis plot/.append style={mark size=\marksize},
every axis x label/.append style={opacity=\thisXlabelopacity},
every axis y label/.append style={opacity=\thisYlabelopacity},
every x tick label/.append style={rotate=0},
height=\figheight,
legend cell align={left},
legend style={fill opacity=0.8, draw opacity=1, text opacity=1, draw=lightgray204},
mark size=\marksize,
tick align=outside,
tick pos=left,
title={FFHQ - Motion Deblur},
width=\figwidth,
x grid style={darkgray176},
xlabel={ODE steps},
xmajorgrids,
xmin=-0.2, xmax=26.2,
xtick style={color=black},
xticklabel style={opacity=\thisXticklabelopacity, align=center},
y grid style={darkgray176},
ylabel={PSNR},
ymajorgrids,
ymin=25, ymax=34,
ytick style={color=black}
]
\addplot [semithick, purple, mark=*, mark size=3, mark options={solid}]
table {%
1 30.838
2 31.544
5 31.806
10 31.583
20 31.265
25 31.122
};
\addlegendentry{psnr}
\ifthenelse{\boolean{NOLEGEND}}{\legend{}}{}
\end{axis}

\end{tikzpicture} \\
\begin{tikzpicture}
\providecommand{\thisYlabelopacity}{1.0}%
\providecommand{\figwidth}{360pt}%
\provideboolean{CLEANTITLE}\ifthenelse{\boolean{CLEANTITLE}}{%
	\pgfplotsset{every axis post/.append style={title = {} }}%
}{}%
\providecommand{\figheight}{310pt}%
\provideboolean{CLEANYAXIS}\ifthenelse{\boolean{CLEANYAXIS}}{%
	\pgfplotsset{every axis post/.append style={ylabel = {} }}%
}{}%
\providecommand{\thisXticklabelopacity}{1.0}%
\provideboolean{CLEANYAXIS}\ifthenelse{\boolean{CLEANYAXIS}}{%
	\pgfplotsset{every axis post/.append style={yticklabels = {} }}%
}{}%
\provideboolean{CLEANXAXIS}\ifthenelse{\boolean{CLEANXAXIS}}{%
	\pgfplotsset{every axis post/.append style={xticklabels = {} }}%
}{}%
\provideboolean{CLEANXAXIS}\ifthenelse{\boolean{CLEANXAXIS}}{%
	\pgfplotsset{every axis post/.append style={xlabel = {} }}%
}{}%
\providecommand{\figheight}{2.0in}%
\provideboolean{NOLEGEND}%
\pgfplotsset{compat=1.15}%
\providecommand{\figwidth}{5.7in}%
\provideboolean{CLEARYLABEL}\ifthenelse{\boolean{CLEARYLABEL}}{%
	\pgfplotsset{every axis post/.append style={ylabel = {} }}%
}{}%
\providecommand{\thisXlabelopacity}{1.0}%
\providecommand{\marksize}{1}%
\provideboolean{CLEARXLABEL}\ifthenelse{\boolean{CLEARXLABEL}}{%
	\pgfplotsset{every axis post/.append style={xlabel = {} }}%
}{}%

\definecolor{darkgray176}{RGB}{176,176,176}
\definecolor{green}{RGB}{0,128,0}
\definecolor{lightgray204}{RGB}{204,204,204}

\begin{axis}[
axis lines=left,
every axis plot/.append style={mark size=\marksize},
every axis x label/.append style={opacity=\thisXlabelopacity},
every axis y label/.append style={opacity=\thisYlabelopacity},
every x tick label/.append style={rotate=0},
height=\figheight,
legend cell align={left},
legend style={fill opacity=0.8, draw opacity=1, text opacity=1, draw=lightgray204},
mark size=\marksize,
tick align=outside,
tick pos=left,
title={FFHQ - Motion Deblur},
width=\figwidth,
x grid style={darkgray176},
xlabel={ODE steps},
xmajorgrids,
xmin=-0.2, xmax=26.2,
xtick style={color=black},
xticklabel style={opacity=\thisXticklabelopacity, align=center},
y grid style={darkgray176},
ylabel={LPIPS},
ymajorgrids,
ymin=0.1196, ymax=0.1724,
ytick style={color=black}
]
\addplot [semithick, green, mark=*, mark size=3, mark options={solid}]
table {%
1 0.17
2 0.146
5 0.124
10 0.122
20 0.122
25 0.123
};
\addlegendentry{lpips}
\ifthenelse{\boolean{NOLEGEND}}{\legend{}}{}
\end{axis}

\end{tikzpicture} \\
    \end{tabular}%

    \captioning{Ablation study: number of ODE steps.}{
    }
    \label{fig:ablation}
\end{figure}

\section{Conclusions}  
\label{sec:conclusions}  
Score-based ODE samplers benefit from low discretization error, enabling fast convergence and high-quality reconstructions, while SDE samplers inject stochasticity at each step to contract accumulated errors. Restart sampling combines both advantages by interleaving short ODE trajectories with large noise injections, producing higher-quality samples.

We extend this concept to posterior sampling in RePS by running multiple conditioned ODE steps interleaved with restart operations. We propose a simple and task-agnostic restart strategy that is easy to implement. This design achieves both faster convergence and enhanced sample fidelity compared to existing baselines.

Our experiments show that RePS consistently produces better-quality samples across a diverse set of tasks, demonstrating that the restart framework offers an efficient and effective way to improve sampling quality.

{
    \small
    \bibliographystyle{ieeenat_fullname}
    \bibliography{%
        diffusion
    }
}

\clearpage
\setcounter{page}{1}
\maketitlesupplementary

\onecolumn


\section{Additional Methods}
\subsection{Posterior score} 
\label{sec:posterior-score}
We can write posterior score, $\nabla_{\argltnts{t}} \log \generposterior{latent=\generltnts{t},patent=\dataobsvs} $, in terms of  $\xpct{}{\Generltnts{0} \middle|\generltnts{t},\dataobsvs} $ following the simple derivation given below.  
\begin{align} 
    \nabla_{\argltnts{t}}
    \log \generposterior{latent=\generltnts{t},patent=\dataobsvs} 
        &=
    \frac{\nabla_{\argltnts{t}}\generposterior{latent=\generltnts{t},patent=\dataobsvs} }{\generposterior{latent=\generltnts{t},patent=\dataobsvs} } \\
        &=
    \def\integrand#1 {%
        \assignkeys{distributions, gener, adjust, #1}%
	    \generdistrvar\left(
            \generltnts{t},\llatent\middle|\patent\paramdisplay
        \right)	
	}
    \frac{1}{\generposterior{latent=\generltnts{t},patent=\dataobsvs} } \nabla_{\argltnts{t}}
    \cmarginalize{llatent/\generltnts{0}}{\integrand}   
\end{align}
Using independence between $\generltnts{t}$ and $\dataobsvs$ conditioned on $\generltnts{0}$ and taking $\nabla_{\argltnts{t}}$ inside the integral,
\begin{equation}
= \frac{1}{\generposterior{latent=\generltnts{t},patent=\dataobsvs} } 
\int_{\generltnts{0}}
\generposterior{latent=\generltnts{0},patent=\dataobsvs} \nabla_{\argltnts{t}} \generposterior{latent=\generltnts{t},patent=\generltnts{0}} \dfrntl{\generltnts{0}}
\end{equation}
Using $  \nabla_{\argltnts{t}} \generposterior{latent=\generltnts{t},patent=\generltnts{0}} = \nabla_{\argltnts{t}} \log \generposterior{latent=\generltnts{t},patent=\generltnts{0}} \generposterior{latent=\generltnts{t},patent=\generltnts{0}} $
we obtain
\begin{align}
\nabla_{\argltnts{t}} \log \generposterior{latent=\generltnts{t},patent=\dataobsvs} 
    &= \int_{\generltnts{0}} 
    \frac{\generposterior{latent=\generltnts{0},patent=\dataobsvs} \generposterior{latent=\generltnts{t},patent=\generltnts{0}} }                    {\generposterior{latent=\generltnts{t},patent=\dataobsvs} } 
    \nabla_{\argltnts{t}} \log \generposterior{latent=\generltnts{t},patent=\generltnts{0}}  \dfrntl{\generltnts{0}} \\
    &= \int_{\generltnts{0}} 
    \generposterior{latent=\generltnts{0},patent={\generltnts{t}, \dataobsvs}}  
    \nabla_{\argltnts{t}} \log \generposterior{latent=\generltnts{t},patent=\generltnts{0}} 
    \dfrntl{\generltnts{0}} \\
    &= \int_{\generltnts{0}}
    \generposterior{latent=\generltnts{0},patent={\generltnts{t}, \dataobsvs}}  
    \bigl(\frac{\generltnts{0} - \generltnts{t}}{\sigma^2(t)}  \bigr)
    \dfrntl{\generltnts{0}} \\
    &= \frac{\xpct{}{\Generltnts{0} \middle|\generltnts{t},\dataobsvs}  - \generltnts{t}}{\sigma^2(t)} 
\end{align}

\subsection{Conditioned SDE}

In general, the forward diffusion process is written in the form:
\begin{equation}
  \dfrntl{\generltnts{t}}
    = \mathbf{f}(\generltnts{t},t)\dfrntl{t}
    + g(t)\;\dfrntl{\wieners{t}},
\end{equation}
where $\generltnts{t}\in\mathbb R^d$ denotes the state at time $t$, $\wieners{t}$ is a standard Wiener process, \(\mathbf{f}(\cdot,t)\) is the drift vector field, and $g(t)$ is the scalar diffusion coefficient.

The corresponding reverse‑time stochastic differential equation is given by:
\begin{equation} \label{eq:general-reverse-sde}
  \dfrntl{\generltnts{t}}
    = \Bigl(
      \mathbf{f}(\generltnts{t},t) - 
      g^2(t) \nabla_{\argltnts{t}}\log 
        \generprior{index=t} 
        \Bigr) \dfrntl{t}
    + g(t) \dfrntl{\revwieners{t}},
\end{equation}
where $\generprior{index=t} $ is the marginal density of $\generltnts{t} $ and $\revwieners{t} $ denotes a reverse‑time Wiener process.
\cref{eq:general-reverse-sde} is solved backwards in time. 

Using Euler–Maruyama method, we can write the one step update as:
\begin{equation}
\generltnts{t-\Delta t}
    = \generltnts{t} - 
    \Bigl(
      \mathbf{f}(\generltnts{t},t) - 
      g^2(t)) \nabla_{\argltnts{t}} \log 
        \generprior{index=t} 
        \Bigr) \Delta t
    + g(t) \sqrt{\Delta t}\, \rv{z},
\end{equation}
where $\rv{z} \sim \nrml{\vect{0}}{\mat{I}}$  and $\Delta t>0$.
For the specific SDE given by \cref{eqn:reverse-sde}, and with $\sigma(t) = t$ and $\dot\sigma(t) = 1$, we can write this as:
\begin{equation}
\generltnts{t-\Delta t}
    = \generltnts{t} + 2\sigma(t)
    \Bigl( \sigma(t) - \sigma(t-\Delta t) \Bigr)
        \nabla_{\argltnts{t}} \log \generprior{index=t} 
    + \sqrt{2\sigma(t)
    \Bigl( \sigma(t) - \sigma(t-\Delta t) \Bigr)}\, \rv{z},
\end{equation}
In order to use this SDE to sample from the posterior distribution $\generposterior{latent=\generltnts{0},patent=\dataobsvs} $, we can write this in terms of posterior score, $\nabla_{\argltnts{t}} \log \generposterior{latent=\generltnts{t},patent=\dataobsvs} $ as:
\begin{equation}
\generltnts{t-\Delta t}
    = \generltnts{t} + 2\sigma(t)
    \Bigl( \sigma(t) - \sigma(t-\Delta t) \Bigr)
        \nabla_{\argltnts{t}} \log \generposterior{latent=\generltnts{t},patent=\dataobsvs} 
    + \sqrt{2\sigma(t)
    \Bigl( \sigma(t) - \sigma(t-\Delta t) \Bigr)}\, \rv{z},
\end{equation}
From \cref{sec:posterior-score}, we can write:
\begin{equation}
\generltnts{t-\Delta t}
    = \generltnts{t} + \frac{2
        \Bigl( \sigma(t) - \sigma(t-\Delta t) \Bigr)
    }{\sigma(t)}  
    \Bigl(
    \xpct{}{\Generltnts{0} \middle|\generltnts{t},\dataobsvs}  - \generltnts{t} 
    \Bigr)
    + \sqrt{2\sigma(t)
        \Bigl( \sigma(t) - \sigma(t-\Delta t) \Bigr)}\, \rv{z},
\end{equation}
We can approximate the mean of the posterior $\generposterior{latent=\generltnts{0},patent={\generltnts{t},\dataobsvs}} $ by the MAP estimate given by \cref{eqn:map-optimization} to get the conditioned SDE.

\section{Experimental Details}
\label{sec:exp-details}
We follow \citet{zhang_improving_2025} for the implementation of inverse problems and use fixed random seeds to ensure fair comparison and reproducible results.

\subsection{ODE solver}
To ensure a fair comparison with the baselines, we used pretrained models. Both baselines were trained with the DDPM objective, which corresponds to the VP–SDE. \citet{karras_elucidating_2022}, however, showed that the sampler can be decoupled from the training strategy, and used a VE–SDE (see \cref{eqn:reverse-ode}) to generate samples. When evaluating scores, a network trained under one parametrization requires appropriate preconditioning to be used with a different sampler; we follow the preconditioning procedure from \citet{karras_elucidating_2022} precisely.

\subsection{Hyper-parameters}
We simulate the conditioned ODE using the Euler method from $\sigma_{\max}=100$ (or from restart levels given by $\sigma_{r}$) to $\sigma_{0}=0.01$ in $10$ steps. The ODE noise schedule follows the polynomial schedule given in \cref{eqn:sigma-schedule} with $\rho_{\mathrm{ode}}=7$. The restart schedule is annealed using a polynomial schedule with $\rho=15$, from $\sigma_{\mathrm{restart}}$ (tuned per task) to $\sigma_{\min}=0.1$. In addition to $\sigma_{\mathrm{restart}}$, the parameters we tune are the number of gradient updates per conditioning step $N$, the learning rate $\eta$, and the relative scale $\lambda$. For gradient-based optimization we use the Adam optimizer, with the learning rate as specified above and all other Adam parameters left at their default values.

Tuning parameters for each task are listed in \cref{tab:params}. Since we always use a fixed number of ODE steps (i.e., $10$), we vary the number of restart levels to obtain results for different NFE budgets. For example, for RePS-1k we use $N_{\mathrm{restart}}=100$, and for RePS-4k we use $N_{\mathrm{restart}}=400$. The full algorithm is given in \cref{alg:reps}.

\tuningParams
\repsAlgo
\qualitative
\multiruns

\section{Additional Results}
\label{sec:add-results}
We visualize qualitative results for selected tasks in \cref{fig:qualitative}, which show that RePS generates fine details consistent with the measurement signal.

We also provide multiple-run examples for phase retrieval and inpaint-box in \cref{fig:multi-runs}, demonstrating reliable reconstruction of the target image in the challenging phase-retrieval task and diverse, plausible completions (e.g., smiles and other facial features) for the masked region in the inpainting examples. 

Additional visualizations are provided in \cref{fig:qualitative-ffhq1,fig:qualitative-ffhq2,fig:qualitative-ffhq3,fig:qualitative-ffhq4,fig:phase-retrieval-multiruns,fig:inpaint-box-multiruns}.

\subsection{Comparison to DAPS}
For a detailed comparison with DAPS, we present results for both RePS and DAPS across different Neural Function Evaluation (NFE) budgets. Figures \cref{fig:linear-plots-nfe-FFHQ} and \cref{fig:non-linear-plots-nfe-FFHQ} show all four evaluation metrics for different NFEs on FFHQ; analogous results for ImageNet appear in \cref{fig:linear-plots-nfe-Imagenet,fig:non-linear-plots-nfe-Imagenet}.

Because algorithms with the same NFE count can still have different run times, we also report results as a function of wall-clock running time per image measured on an NVIDIA A100-40GB. These results are shown in \cref{fig:linear-plots-time-FFHQ,fig:non-linear-plots-time-FFHQ} for FFHQ and in \cref{fig:linear-plots-time-Imagenet,fig:non-linear-plots-time-Imagenet} for ImageNet.

As noted in the main paper, the results obtained by running the official DAPS code differ slightly from those reported in the DAPS paper. Therefore, all quantitative comparisons in this work use results produced by running the official DAPS implementation. For completeness, we also reproduce the original results reported in the DAPS paper in \cref{tab:supp_results_daps}.

\newpage


\begin{figure*}[t]
  \centering
    \newcommand{\figwidth}{0.22\linewidth}%
    \newcommand{\figheight}{0.22\linewidth}%
    \newcommand{\marksize}{2.5}%
    \setlength{\tabcolsep}{0pt}%
    \provideboolean{NOLEGEND}\setboolean{NOLEGEND}{true}%
    \provideboolean{CLEARYLABEL}\setboolean{CLEARYLABEL}{true}%
    \provideboolean{CLEANTITLE}\setboolean{CLEANTITLE}{true}%
    \provideboolean{CLEANXAXIS}\setboolean{CLEANXAXIS}{true}%
    \newcommand{\DATASET}{ffhq}
    \newcommand{\QUANTITY}{nfe}
    \newcommand{\RESULTTYPE}{reps-daps}

        \newcommand{\tikzsubdir}{\tikzdir/\RESULTTYPE}%
    \begin{tabular}{@{}ccccc@{}}

        & PSNR ($\uparrow$) & SSIM ($\uparrow$) & LPIPS ($\downarrow$) & FID ($\downarrow$) \\    
        \provideboolean{CLEANXAXIS}\setboolean{CLEANXAXIS}{true}%
        \raisebox{0.08\linewidth}{super-resolution} &
        \input{tikzpics/\RESULTTYPE/\RESULTTYPE-\DATASET-super-resolution-psnr-\QUANTITY} &
        \input{tikzpics/\RESULTTYPE/\RESULTTYPE-\DATASET-super-resolution-ssim-\QUANTITY} &
        \input{tikzpics/\RESULTTYPE/\RESULTTYPE-\DATASET-super-resolution-lpips-\QUANTITY} &
        \input{tikzpics/\RESULTTYPE/\RESULTTYPE-\DATASET-super-resolution-fid-\QUANTITY} \\ 
        \raisebox{0.08\linewidth}{inpaint-box} &
        \input{tikzpics/\RESULTTYPE/\RESULTTYPE-\DATASET-inpaint-box-psnr-\QUANTITY} &
        \input{tikzpics/\RESULTTYPE/\RESULTTYPE-\DATASET-inpaint-box-ssim-\QUANTITY} &
        \input{tikzpics/\RESULTTYPE/\RESULTTYPE-\DATASET-inpaint-box-lpips-\QUANTITY} &
        \input{tikzpics/\RESULTTYPE/\RESULTTYPE-\DATASET-inpaint-box-fid-\QUANTITY} \\ 
        \raisebox{0.08\linewidth}{inpaint-random} &
        \input{tikzpics/\RESULTTYPE/\RESULTTYPE-\DATASET-inpaint-random-psnr-\QUANTITY} &
        \input{tikzpics/\RESULTTYPE/\RESULTTYPE-\DATASET-inpaint-random-ssim-\QUANTITY} &
        \input{tikzpics/\RESULTTYPE/\RESULTTYPE-\DATASET-inpaint-random-lpips-\QUANTITY} &
        \input{tikzpics/\RESULTTYPE/\RESULTTYPE-\DATASET-inpaint-random-fid-\QUANTITY} \\ 
        \raisebox{0.08\linewidth}{gaussian-deblur} &
        \input{tikzpics/\RESULTTYPE/\RESULTTYPE-\DATASET-gaussian-deblur-psnr-\QUANTITY} &
        \input{tikzpics/\RESULTTYPE/\RESULTTYPE-\DATASET-gaussian-deblur-ssim-\QUANTITY} &
        \input{tikzpics/\RESULTTYPE/\RESULTTYPE-\DATASET-gaussian-deblur-lpips-\QUANTITY} &
        \input{tikzpics/\RESULTTYPE/\RESULTTYPE-\DATASET-gaussian-deblur-fid-\QUANTITY} \\ 
        \raisebox{0.13\linewidth}{motion-deblur} &
        \provideboolean{CLEANXAXIS}\setboolean{CLEANXAXIS}{false}%
        \input{tikzpics/\RESULTTYPE/\RESULTTYPE-\DATASET-motion-deblur-psnr-\QUANTITY} &
        \provideboolean{CLEANXAXIS}\setboolean{CLEANXAXIS}{false}%
        \input{tikzpics/\RESULTTYPE/\RESULTTYPE-\DATASET-motion-deblur-ssim-\QUANTITY} &
        \provideboolean{CLEANXAXIS}\setboolean{CLEANXAXIS}{false}%
        \input{tikzpics/\RESULTTYPE/\RESULTTYPE-\DATASET-motion-deblur-lpips-\QUANTITY} &
        \provideboolean{CLEANXAXIS}\setboolean{CLEANXAXIS}{false}%
        \input{tikzpics/\RESULTTYPE/\RESULTTYPE-\DATASET-motion-deblur-fid-\QUANTITY} \\ 
    \end{tabular}%

    \captioning{Quantitative results vs NFEs for linear problems on FFHQ.}{Figure shows all four metrics as a function of NFEs for RePS (orange) and DAPS (blue) for all linear problems on FFHQ dataset.}
    \label{fig:linear-plots-nfe-FFHQ}
\end{figure*}

\begin{figure*}[t]
  \centering
    \newcommand{\figwidth}{0.22\linewidth}%
    \newcommand{\figheight}{0.22\linewidth}%
    \newcommand{\marksize}{2.5}%
    \setlength{\tabcolsep}{0pt}%
    \provideboolean{NOLEGEND}\setboolean{NOLEGEND}{true}%
    \provideboolean{CLEARYLABEL}\setboolean{CLEARYLABEL}{true}%
    \provideboolean{CLEANTITLE}\setboolean{CLEANTITLE}{true}%
    \provideboolean{CLEANXAXIS}\setboolean{CLEANXAXIS}{true}%
    
    \newcommand{\DATASET}{ffhq}
    \newcommand{\QUANTITY}{nfe}
    \newcommand{\RESULTTYPE}{reps-daps}

        \newcommand{\tikzsubdir}{\tikzdir/\RESULTTYPE}%
    \begin{tabular}{@{}ccccc@{}}

        & PSNR ($\uparrow$) & SSIM ($\uparrow$) & LPIPS ($\downarrow$) & FID ($\downarrow$) \\   
    
        \provideboolean{CLEANXAXIS}\setboolean{CLEANXAXIS}{true}%
        \raisebox{0.08\linewidth}{phase-retrieval} &
        \input{tikzpics/\RESULTTYPE/\RESULTTYPE-\DATASET-phase-retrieval-psnr-\QUANTITY} &
        \input{tikzpics/\RESULTTYPE/\RESULTTYPE-\DATASET-phase-retrieval-ssim-\QUANTITY} &
        \input{tikzpics/\RESULTTYPE/\RESULTTYPE-\DATASET-phase-retrieval-lpips-\QUANTITY} &
        \input{tikzpics/\RESULTTYPE/\RESULTTYPE-\DATASET-phase-retrieval-fid-\QUANTITY} \\ 
        \raisebox{0.08\linewidth}{nonlinear-deblur} &
        \input{tikzpics/\RESULTTYPE/\RESULTTYPE-\DATASET-nonlinear-deblur-psnr-\QUANTITY} &
        \input{tikzpics/\RESULTTYPE/\RESULTTYPE-\DATASET-nonlinear-deblur-ssim-\QUANTITY} &
        \input{tikzpics/\RESULTTYPE/\RESULTTYPE-\DATASET-nonlinear-deblur-lpips-\QUANTITY} &
        \input{tikzpics/\RESULTTYPE/\RESULTTYPE-\DATASET-nonlinear-deblur-fid-\QUANTITY} \\ 
        \raisebox{0.13\linewidth}{hdr} &
        \provideboolean{CLEANXAXIS}\setboolean{CLEANXAXIS}{false}%
        \input{tikzpics/\RESULTTYPE/\RESULTTYPE-\DATASET-hdr-psnr-\QUANTITY} &
        \provideboolean{CLEANXAXIS}\setboolean{CLEANXAXIS}{false}%
        \input{tikzpics/\RESULTTYPE/\RESULTTYPE-\DATASET-hdr-ssim-\QUANTITY} &
        \provideboolean{CLEANXAXIS}\setboolean{CLEANXAXIS}{false}%
        \input{tikzpics/\RESULTTYPE/\RESULTTYPE-\DATASET-hdr-lpips-\QUANTITY} &
        \provideboolean{CLEANXAXIS}\setboolean{CLEANXAXIS}{false}%
        \input{tikzpics/\RESULTTYPE/\RESULTTYPE-\DATASET-hdr-fid-\QUANTITY} \\ 
    \end{tabular}%

    \captioning{Quantitative results vs NFEs for non-linear problems on FFHQ.}{Figure shows all four metrics as a function of NFEs for RePS (orange) and DAPS (blue) for all the non-linear problems on FFHQ dataset.}
    \label{fig:non-linear-plots-nfe-FFHQ}
\end{figure*}


\begin{figure*}[t]
  \centering
    \newcommand{\figwidth}{0.22\linewidth}%
    \newcommand{\figheight}{0.22\linewidth}%
    \newcommand{\marksize}{2.5}%
    \setlength{\tabcolsep}{0pt}%
    \provideboolean{NOLEGEND}\setboolean{NOLEGEND}{true}%
    \provideboolean{CLEARYLABEL}\setboolean{CLEARYLABEL}{true}%
    \provideboolean{CLEANTITLE}\setboolean{CLEANTITLE}{true}%
    \provideboolean{CLEANXAXIS}\setboolean{CLEANXAXIS}{true}%
    
    \newcommand{\DATASET}{ffhq}
    \newcommand{\QUANTITY}{time}
    \newcommand{\RESULTTYPE}{reps-daps}

        \newcommand{\tikzsubdir}{\tikzdir/\RESULTTYPE}%
    \begin{tabular}{@{}ccccc@{}}

        & PSNR ($\uparrow$) & SSIM ($\uparrow$) & LPIPS ($\downarrow$) & FID ($\downarrow$) \\    
        \provideboolean{CLEANXAXIS}\setboolean{CLEANXAXIS}{true}%
        \raisebox{0.08\linewidth}{super-resolution} &
        \input{tikzpics/\RESULTTYPE/\RESULTTYPE-\DATASET-super-resolution-psnr-\QUANTITY} &
        \input{tikzpics/\RESULTTYPE/\RESULTTYPE-\DATASET-super-resolution-ssim-\QUANTITY} &
        \input{tikzpics/\RESULTTYPE/\RESULTTYPE-\DATASET-super-resolution-lpips-\QUANTITY} &
        \input{tikzpics/\RESULTTYPE/\RESULTTYPE-\DATASET-super-resolution-fid-\QUANTITY} \\ 
        \raisebox{0.08\linewidth}{inpaint-box} &
        \input{tikzpics/\RESULTTYPE/\RESULTTYPE-\DATASET-inpaint-box-psnr-\QUANTITY} &
        \input{tikzpics/\RESULTTYPE/\RESULTTYPE-\DATASET-inpaint-box-ssim-\QUANTITY} &
        \input{tikzpics/\RESULTTYPE/\RESULTTYPE-\DATASET-inpaint-box-lpips-\QUANTITY} &
        \input{tikzpics/\RESULTTYPE/\RESULTTYPE-\DATASET-inpaint-box-fid-\QUANTITY} \\ 
        \raisebox{0.08\linewidth}{inpaint-random} &
        \input{tikzpics/\RESULTTYPE/\RESULTTYPE-\DATASET-inpaint-random-psnr-\QUANTITY} &
        \input{tikzpics/\RESULTTYPE/\RESULTTYPE-\DATASET-inpaint-random-ssim-\QUANTITY} &
        \input{tikzpics/\RESULTTYPE/\RESULTTYPE-\DATASET-inpaint-random-lpips-\QUANTITY} &
        \input{tikzpics/\RESULTTYPE/\RESULTTYPE-\DATASET-inpaint-random-fid-\QUANTITY} \\ 
        \raisebox{0.08\linewidth}{gaussian-deblur} &
        \input{tikzpics/\RESULTTYPE/\RESULTTYPE-\DATASET-gaussian-deblur-psnr-\QUANTITY} &
        \input{tikzpics/\RESULTTYPE/\RESULTTYPE-\DATASET-gaussian-deblur-ssim-\QUANTITY} &
        \input{tikzpics/\RESULTTYPE/\RESULTTYPE-\DATASET-gaussian-deblur-lpips-\QUANTITY} &
        \input{tikzpics/\RESULTTYPE/\RESULTTYPE-\DATASET-gaussian-deblur-fid-\QUANTITY} \\ 
        \raisebox{0.13\linewidth}{motion-deblur} &
        \provideboolean{CLEANXAXIS}\setboolean{CLEANXAXIS}{false}%
        \input{tikzpics/\RESULTTYPE/\RESULTTYPE-\DATASET-motion-deblur-psnr-\QUANTITY} &
        \provideboolean{CLEANXAXIS}\setboolean{CLEANXAXIS}{false}%
        \input{tikzpics/\RESULTTYPE/\RESULTTYPE-\DATASET-motion-deblur-ssim-\QUANTITY} &
        \provideboolean{CLEANXAXIS}\setboolean{CLEANXAXIS}{false}%
        \input{tikzpics/\RESULTTYPE/\RESULTTYPE-\DATASET-motion-deblur-lpips-\QUANTITY} &
        \provideboolean{CLEANXAXIS}\setboolean{CLEANXAXIS}{false}%
        \input{tikzpics/\RESULTTYPE/\RESULTTYPE-\DATASET-motion-deblur-fid-\QUANTITY} \\ 
    \end{tabular}%

    \captioning{Quantitative results vs running time-per-sample for linear problems on FFHQ.}{Figure shows all four metrics as a function of running time-per-sample (seconds) for RePS (orange) and DAPS (blue) for all the linear problems on FFHQ dataset. All samples were generated on an NVIDIA A100-40GB GPU.}
    \label{fig:linear-plots-time-FFHQ}
\end{figure*}

\begin{figure*}[t]
  \centering
    \newcommand{\figwidth}{0.22\linewidth}%
    \newcommand{\figheight}{0.22\linewidth}%
    \newcommand{\marksize}{2.5}%
    \setlength{\tabcolsep}{0pt}%
    \provideboolean{NOLEGEND}\setboolean{NOLEGEND}{true}%
    \provideboolean{CLEARYLABEL}\setboolean{CLEARYLABEL}{true}%
    \provideboolean{CLEANTITLE}\setboolean{CLEANTITLE}{true}%
    \provideboolean{CLEANXAXIS}\setboolean{CLEANXAXIS}{true}%
    
    \newcommand{\DATASET}{ffhq}
    \newcommand{\QUANTITY}{time}
    \newcommand{\RESULTTYPE}{reps-daps}

        \newcommand{\tikzsubdir}{\tikzdir/\RESULTTYPE}%
    \begin{tabular}{@{}ccccc@{}}

        & PSNR ($\uparrow$) & SSIM ($\uparrow$) & LPIPS ($\downarrow$) & FID ($\downarrow$) \\   
    
        \provideboolean{CLEANXAXIS}\setboolean{CLEANXAXIS}{true}%
        \raisebox{0.08\linewidth}{phase-retrieval} &
        \input{tikzpics/\RESULTTYPE/\RESULTTYPE-\DATASET-phase-retrieval-psnr-\QUANTITY} &
        \input{tikzpics/\RESULTTYPE/\RESULTTYPE-\DATASET-phase-retrieval-ssim-\QUANTITY} &
        \input{tikzpics/\RESULTTYPE/\RESULTTYPE-\DATASET-phase-retrieval-lpips-\QUANTITY} &
        \input{tikzpics/\RESULTTYPE/\RESULTTYPE-\DATASET-phase-retrieval-fid-\QUANTITY} \\ 
        \raisebox{0.08\linewidth}{nonlinear-deblur} &
        \input{tikzpics/\RESULTTYPE/\RESULTTYPE-\DATASET-nonlinear-deblur-psnr-\QUANTITY} &
        \input{tikzpics/\RESULTTYPE/\RESULTTYPE-\DATASET-nonlinear-deblur-ssim-\QUANTITY} &
        \input{tikzpics/\RESULTTYPE/\RESULTTYPE-\DATASET-nonlinear-deblur-lpips-\QUANTITY} &
        \input{tikzpics/\RESULTTYPE/\RESULTTYPE-\DATASET-nonlinear-deblur-fid-\QUANTITY} \\ 
        \raisebox{0.13\linewidth}{hdr} &
        \provideboolean{CLEANXAXIS}\setboolean{CLEANXAXIS}{false}%
        \input{tikzpics/\RESULTTYPE/\RESULTTYPE-\DATASET-hdr-psnr-\QUANTITY} &
        \provideboolean{CLEANXAXIS}\setboolean{CLEANXAXIS}{false}%
        \input{tikzpics/\RESULTTYPE/\RESULTTYPE-\DATASET-hdr-ssim-\QUANTITY} &
        \provideboolean{CLEANXAXIS}\setboolean{CLEANXAXIS}{false}%
        \input{tikzpics/\RESULTTYPE/\RESULTTYPE-\DATASET-hdr-lpips-\QUANTITY} &
        \provideboolean{CLEANXAXIS}\setboolean{CLEANXAXIS}{false}%
        \input{tikzpics/\RESULTTYPE/\RESULTTYPE-\DATASET-hdr-fid-\QUANTITY} \\ 
    \end{tabular}%

    \captioning{Quantitative results vs running time-per-sample for non-linear problems on FFHQ.}{Figure shows all four metrics as a function of running time-per-sample (seconds) for RePS (orange) and DAPS (blue) for all the non-linear problems on FFHQ dataset. All samples were generated on an NVIDIA A100-40GB GPU.}
    \label{fig:non-linear-plots-time-FFHQ}
\end{figure*}


\begin{figure*}[t]
  \centering
    \newcommand{\figwidth}{0.22\linewidth}%
    \newcommand{\figheight}{0.22\linewidth}%
    \newcommand{\marksize}{2.5}%
    \setlength{\tabcolsep}{0pt}%
    \provideboolean{NOLEGEND}\setboolean{NOLEGEND}{true}%
    \provideboolean{CLEARYLABEL}\setboolean{CLEARYLABEL}{true}%
    \provideboolean{CLEANTITLE}\setboolean{CLEANTITLE}{true}%
    \provideboolean{CLEANXAXIS}\setboolean{CLEANXAXIS}{true}%
    \newcommand{\DATASET}{imagenet}
    \newcommand{\QUANTITY}{nfe}
    \newcommand{\RESULTTYPE}{reps-daps}

        \newcommand{\tikzsubdir}{\tikzdir/\RESULTTYPE}%
    \begin{tabular}{@{}ccccc@{}}

        & PSNR ($\uparrow$) & SSIM ($\uparrow$) & LPIPS ($\downarrow$) & FID ($\downarrow$) \\    
        \provideboolean{CLEANXAXIS}\setboolean{CLEANXAXIS}{true}%
        \raisebox{0.08\linewidth}{super-resolution} &
        \input{tikzpics/\RESULTTYPE/\RESULTTYPE-\DATASET-super-resolution-psnr-\QUANTITY} &
        \input{tikzpics/\RESULTTYPE/\RESULTTYPE-\DATASET-super-resolution-ssim-\QUANTITY} &
        \input{tikzpics/\RESULTTYPE/\RESULTTYPE-\DATASET-super-resolution-lpips-\QUANTITY} &
        \input{tikzpics/\RESULTTYPE/\RESULTTYPE-\DATASET-super-resolution-fid-\QUANTITY} \\ 
        \raisebox{0.08\linewidth}{inpaint-box} &
        \input{tikzpics/\RESULTTYPE/\RESULTTYPE-\DATASET-inpaint-box-psnr-\QUANTITY} &
        \input{tikzpics/\RESULTTYPE/\RESULTTYPE-\DATASET-inpaint-box-ssim-\QUANTITY} &
        \input{tikzpics/\RESULTTYPE/\RESULTTYPE-\DATASET-inpaint-box-lpips-\QUANTITY} &
        \input{tikzpics/\RESULTTYPE/\RESULTTYPE-\DATASET-inpaint-box-fid-\QUANTITY} \\ 
        \raisebox{0.08\linewidth}{inpaint-random} &
        \input{tikzpics/\RESULTTYPE/\RESULTTYPE-\DATASET-inpaint-random-psnr-\QUANTITY} &
        \input{tikzpics/\RESULTTYPE/\RESULTTYPE-\DATASET-inpaint-random-ssim-\QUANTITY} &
        \input{tikzpics/\RESULTTYPE/\RESULTTYPE-\DATASET-inpaint-random-lpips-\QUANTITY} &
        \input{tikzpics/\RESULTTYPE/\RESULTTYPE-\DATASET-inpaint-random-fid-\QUANTITY} \\ 
        \raisebox{0.08\linewidth}{gaussian-deblur} &
        \input{tikzpics/\RESULTTYPE/\RESULTTYPE-\DATASET-gaussian-deblur-psnr-\QUANTITY} &
        \input{tikzpics/\RESULTTYPE/\RESULTTYPE-\DATASET-gaussian-deblur-ssim-\QUANTITY} &
        \input{tikzpics/\RESULTTYPE/\RESULTTYPE-\DATASET-gaussian-deblur-lpips-\QUANTITY} &
        \input{tikzpics/\RESULTTYPE/\RESULTTYPE-\DATASET-gaussian-deblur-fid-\QUANTITY} \\ 
        \raisebox{0.13\linewidth}{motion-deblur} &
        \provideboolean{CLEANXAXIS}\setboolean{CLEANXAXIS}{false}%
        \input{tikzpics/\RESULTTYPE/\RESULTTYPE-\DATASET-motion-deblur-psnr-\QUANTITY} &
        \provideboolean{CLEANXAXIS}\setboolean{CLEANXAXIS}{false}%
        \input{tikzpics/\RESULTTYPE/\RESULTTYPE-\DATASET-motion-deblur-ssim-\QUANTITY} &
        \provideboolean{CLEANXAXIS}\setboolean{CLEANXAXIS}{false}%
        \input{tikzpics/\RESULTTYPE/\RESULTTYPE-\DATASET-motion-deblur-lpips-\QUANTITY} &
        \provideboolean{CLEANXAXIS}\setboolean{CLEANXAXIS}{false}%
        \input{tikzpics/\RESULTTYPE/\RESULTTYPE-\DATASET-motion-deblur-fid-\QUANTITY} \\ 
    \end{tabular}%

    \captioning{Quantitative results vs NFEs for linear problems on Imagenet.}{Figure shows all four metrics as a function of NFEs for RePS (orange) and DAPS (blue) for all linear problems on Imagenet dataset.}
    \label{fig:linear-plots-nfe-Imagenet}
\end{figure*}

\begin{figure*}[t]
  \centering
    \newcommand{\figwidth}{0.22\linewidth}%
    \newcommand{\figheight}{0.22\linewidth}%
    \newcommand{\marksize}{2.5}%
    \setlength{\tabcolsep}{0pt}%
    \provideboolean{NOLEGEND}\setboolean{NOLEGEND}{true}%
    \provideboolean{CLEARYLABEL}\setboolean{CLEARYLABEL}{true}%
    \provideboolean{CLEANTITLE}\setboolean{CLEANTITLE}{true}%
    \provideboolean{CLEANXAXIS}\setboolean{CLEANXAXIS}{true}%
    
    \newcommand{\DATASET}{imagenet}
    \newcommand{\QUANTITY}{nfe}
    \newcommand{\RESULTTYPE}{reps-daps}

        \newcommand{\tikzsubdir}{\tikzdir/\RESULTTYPE}%
    \begin{tabular}{@{}ccccc@{}}

        & PSNR ($\uparrow$) & SSIM ($\uparrow$) & LPIPS ($\downarrow$) & FID ($\downarrow$) \\   
    
        \provideboolean{CLEANXAXIS}\setboolean{CLEANXAXIS}{true}%
        \raisebox{0.08\linewidth}{phase-retrieval} &
        \input{tikzpics/\RESULTTYPE/\RESULTTYPE-\DATASET-phase-retrieval-psnr-\QUANTITY} &
        \input{tikzpics/\RESULTTYPE/\RESULTTYPE-\DATASET-phase-retrieval-ssim-\QUANTITY} &
        \input{tikzpics/\RESULTTYPE/\RESULTTYPE-\DATASET-phase-retrieval-lpips-\QUANTITY} &
        \input{tikzpics/\RESULTTYPE/\RESULTTYPE-\DATASET-phase-retrieval-fid-\QUANTITY} \\ 
        \raisebox{0.08\linewidth}{nonlinear-deblur} &
        \input{tikzpics/\RESULTTYPE/\RESULTTYPE-\DATASET-nonlinear-deblur-psnr-\QUANTITY} &
        \input{tikzpics/\RESULTTYPE/\RESULTTYPE-\DATASET-nonlinear-deblur-ssim-\QUANTITY} &
        \input{tikzpics/\RESULTTYPE/\RESULTTYPE-\DATASET-nonlinear-deblur-lpips-\QUANTITY} &
        \input{tikzpics/\RESULTTYPE/\RESULTTYPE-\DATASET-nonlinear-deblur-fid-\QUANTITY} \\ 
        \raisebox{0.13\linewidth}{hdr} &
        \provideboolean{CLEANXAXIS}\setboolean{CLEANXAXIS}{false}%
        \input{tikzpics/\RESULTTYPE/\RESULTTYPE-\DATASET-hdr-psnr-\QUANTITY} &
        \provideboolean{CLEANXAXIS}\setboolean{CLEANXAXIS}{false}%
        \input{tikzpics/\RESULTTYPE/\RESULTTYPE-\DATASET-hdr-ssim-\QUANTITY} &
        \provideboolean{CLEANXAXIS}\setboolean{CLEANXAXIS}{false}%
        \input{tikzpics/\RESULTTYPE/\RESULTTYPE-\DATASET-hdr-lpips-\QUANTITY} &
        \provideboolean{CLEANXAXIS}\setboolean{CLEANXAXIS}{false}%
        \input{tikzpics/\RESULTTYPE/\RESULTTYPE-\DATASET-hdr-fid-\QUANTITY} \\ 
    \end{tabular}%

    \captioning{Quantitative results vs NFEs for non-linear problems on Imagenet.}{Figure shows all four metrics as a function of NFEs for RePS (orange) and DAPS (blue) for all the non-linear problems on Imagenet dataset.}
    \label{fig:non-linear-plots-nfe-Imagenet}
\end{figure*}


\begin{figure*}[t]
  \centering
    \newcommand{\figwidth}{0.22\linewidth}%
    \newcommand{\figheight}{0.22\linewidth}%
    \newcommand{\marksize}{2.5}%
    \setlength{\tabcolsep}{0pt}%
    \provideboolean{NOLEGEND}\setboolean{NOLEGEND}{true}%
    \provideboolean{CLEARYLABEL}\setboolean{CLEARYLABEL}{true}%
    \provideboolean{CLEANTITLE}\setboolean{CLEANTITLE}{true}%
    \provideboolean{CLEANXAXIS}\setboolean{CLEANXAXIS}{true}%
    
    \newcommand{\DATASET}{imagenet}
    \newcommand{\QUANTITY}{time}
    \newcommand{\RESULTTYPE}{reps-daps}

        \newcommand{\tikzsubdir}{\tikzdir/\RESULTTYPE}%
    \begin{tabular}{@{}ccccc@{}}

        & PSNR ($\uparrow$) & SSIM ($\uparrow$) & LPIPS ($\downarrow$) & FID ($\downarrow$) \\    
        \provideboolean{CLEANXAXIS}\setboolean{CLEANXAXIS}{true}%
        \raisebox{0.08\linewidth}{super-resolution} &
        \input{tikzpics/\RESULTTYPE/\RESULTTYPE-\DATASET-super-resolution-psnr-\QUANTITY} &
        \input{tikzpics/\RESULTTYPE/\RESULTTYPE-\DATASET-super-resolution-ssim-\QUANTITY} &
        \input{tikzpics/\RESULTTYPE/\RESULTTYPE-\DATASET-super-resolution-lpips-\QUANTITY} &
        \input{tikzpics/\RESULTTYPE/\RESULTTYPE-\DATASET-super-resolution-fid-\QUANTITY} \\ 
        \raisebox{0.08\linewidth}{inpaint-box} &
        \input{tikzpics/\RESULTTYPE/\RESULTTYPE-\DATASET-inpaint-box-psnr-\QUANTITY} &
        \input{tikzpics/\RESULTTYPE/\RESULTTYPE-\DATASET-inpaint-box-ssim-\QUANTITY} &
        \input{tikzpics/\RESULTTYPE/\RESULTTYPE-\DATASET-inpaint-box-lpips-\QUANTITY} &
        \input{tikzpics/\RESULTTYPE/\RESULTTYPE-\DATASET-inpaint-box-fid-\QUANTITY} \\ 
        \raisebox{0.08\linewidth}{inpaint-random} &
        \input{tikzpics/\RESULTTYPE/\RESULTTYPE-\DATASET-inpaint-random-psnr-\QUANTITY} &
        \input{tikzpics/\RESULTTYPE/\RESULTTYPE-\DATASET-inpaint-random-ssim-\QUANTITY} &
        \input{tikzpics/\RESULTTYPE/\RESULTTYPE-\DATASET-inpaint-random-lpips-\QUANTITY} &
        \input{tikzpics/\RESULTTYPE/\RESULTTYPE-\DATASET-inpaint-random-fid-\QUANTITY} \\ 
        \raisebox{0.08\linewidth}{gaussian-deblur} &
        \input{tikzpics/\RESULTTYPE/\RESULTTYPE-\DATASET-gaussian-deblur-psnr-\QUANTITY} &
        \input{tikzpics/\RESULTTYPE/\RESULTTYPE-\DATASET-gaussian-deblur-ssim-\QUANTITY} &
        \input{tikzpics/\RESULTTYPE/\RESULTTYPE-\DATASET-gaussian-deblur-lpips-\QUANTITY} &
        \input{tikzpics/\RESULTTYPE/\RESULTTYPE-\DATASET-gaussian-deblur-fid-\QUANTITY} \\ 
        \raisebox{0.13\linewidth}{motion-deblur} &
        \provideboolean{CLEANXAXIS}\setboolean{CLEANXAXIS}{false}%
        \input{tikzpics/\RESULTTYPE/\RESULTTYPE-\DATASET-motion-deblur-psnr-\QUANTITY} &
        \provideboolean{CLEANXAXIS}\setboolean{CLEANXAXIS}{false}%
        \input{tikzpics/\RESULTTYPE/\RESULTTYPE-\DATASET-motion-deblur-ssim-\QUANTITY} &
        \provideboolean{CLEANXAXIS}\setboolean{CLEANXAXIS}{false}%
        \input{tikzpics/\RESULTTYPE/\RESULTTYPE-\DATASET-motion-deblur-lpips-\QUANTITY} &
        \provideboolean{CLEANXAXIS}\setboolean{CLEANXAXIS}{false}%
        \input{tikzpics/\RESULTTYPE/\RESULTTYPE-\DATASET-motion-deblur-fid-\QUANTITY} \\ 
    \end{tabular}%

    \captioning{Quantitative results vs running time-per-sample for linear problems on Imagenet.}{Figure shows all four metrics as a function of running time-per-sample (seconds) for RePS (orange) and DAPS (blue) for all the linear problems on Imagenet dataset. All samples were generated on an NVIDIA A100-40GB GPU.}
    \label{fig:linear-plots-time-Imagenet}
\end{figure*}

\begin{figure*}[t]
  \centering
    \newcommand{\figwidth}{0.22\linewidth}%
    \newcommand{\figheight}{0.22\linewidth}%
    \newcommand{\marksize}{2.5}%
    \setlength{\tabcolsep}{0pt}%
    \provideboolean{NOLEGEND}\setboolean{NOLEGEND}{true}%
    \provideboolean{CLEARYLABEL}\setboolean{CLEARYLABEL}{true}%
    \provideboolean{CLEANTITLE}\setboolean{CLEANTITLE}{true}%
    \provideboolean{CLEANXAXIS}\setboolean{CLEANXAXIS}{true}%
    
    \newcommand{\DATASET}{imagenet}
    \newcommand{\QUANTITY}{time}
    \newcommand{\RESULTTYPE}{reps-daps}

        \newcommand{\tikzsubdir}{\tikzdir/\RESULTTYPE}%
    \begin{tabular}{@{}ccccc@{}}

        & PSNR ($\uparrow$) & SSIM ($\uparrow$) & LPIPS ($\downarrow$) & FID ($\downarrow$) \\   
    
        \provideboolean{CLEANXAXIS}\setboolean{CLEANXAXIS}{true}%
        \raisebox{0.08\linewidth}{phase-retrieval} &
        \input{tikzpics/\RESULTTYPE/\RESULTTYPE-\DATASET-phase-retrieval-psnr-\QUANTITY} &
        \input{tikzpics/\RESULTTYPE/\RESULTTYPE-\DATASET-phase-retrieval-ssim-\QUANTITY} &
        \input{tikzpics/\RESULTTYPE/\RESULTTYPE-\DATASET-phase-retrieval-lpips-\QUANTITY} &
        \input{tikzpics/\RESULTTYPE/\RESULTTYPE-\DATASET-phase-retrieval-fid-\QUANTITY} \\ 
        \raisebox{0.08\linewidth}{nonlinear-deblur} &
        \input{tikzpics/\RESULTTYPE/\RESULTTYPE-\DATASET-nonlinear-deblur-psnr-\QUANTITY} &
        \input{tikzpics/\RESULTTYPE/\RESULTTYPE-\DATASET-nonlinear-deblur-ssim-\QUANTITY} &
        \input{tikzpics/\RESULTTYPE/\RESULTTYPE-\DATASET-nonlinear-deblur-lpips-\QUANTITY} &
        \input{tikzpics/\RESULTTYPE/\RESULTTYPE-\DATASET-nonlinear-deblur-fid-\QUANTITY} \\ 
        \raisebox{0.13\linewidth}{hdr} &
        \provideboolean{CLEANXAXIS}\setboolean{CLEANXAXIS}{false}%
        \input{tikzpics/\RESULTTYPE/\RESULTTYPE-\DATASET-hdr-psnr-\QUANTITY} &
        \provideboolean{CLEANXAXIS}\setboolean{CLEANXAXIS}{false}%
        \input{tikzpics/\RESULTTYPE/\RESULTTYPE-\DATASET-hdr-ssim-\QUANTITY} &
        \provideboolean{CLEANXAXIS}\setboolean{CLEANXAXIS}{false}%
        \input{tikzpics/\RESULTTYPE/\RESULTTYPE-\DATASET-hdr-lpips-\QUANTITY} &
        \provideboolean{CLEANXAXIS}\setboolean{CLEANXAXIS}{false}%
        \input{tikzpics/\RESULTTYPE/\RESULTTYPE-\DATASET-hdr-fid-\QUANTITY} \\ 
    \end{tabular}%

    \captioning{Quantitative results vs running time-per-sample for non-linear problems on Imagenet.}{Figure shows all four metrics as a function of running time-per-sample (seconds) for RePS (orange) and DAPS (blue) for all the non-linear problems on Imagenet dataset. All samples were generated on an NVIDIA A100-40GB GPU.}
    \label{fig:non-linear-plots-time-Imagenet}
\end{figure*}

\DAPScodevspaper

\FFHQFiguresFour
\FFHQFiguresOne
\FFHQFiguresTwo
\FFHQFiguresThree

\phaseRetrievalMultiruns
\inpaintBoxMultiruns

\end{document}